\documentclass[hidelinks]{amsart}
\usepackage{float}
\usepackage{mathtools}
\usepackage{xcolor,enumerate}       
 \definecolor{darkgreen}{rgb}{0,0.5,0}

\usepackage[T1]{fontenc}    
\usepackage{hyperref}       
\usepackage{url}            
\usepackage{booktabs}       
\usepackage{amsfonts}       
\usepackage{nicefrac}       
\usepackage{microtype}      
\usepackage{amssymb,latexsym,amsfonts,amsmath}
\usepackage{mathrsfs}
\usepackage{bbm}
\usepackage{graphicx}
\usepackage{color}
\usepackage{mathabx}
\usepackage{algorithm}
\usepackage[noend]{algpseudocode}
\usepackage{bigints}
\usepackage{float}

\usepackage{color}
\usepackage{caption,subfigure, mdframed, tcolorbox, xcolor}
\usepackage{tikz}
\usetikzlibrary{arrows, positioning}
 \definecolor{darkgreen}{rgb}{0,0.5,0}
\tikzset{every node/.style={font=\small}}

\newcommand{\real}{{\mathbb{R}}}

\newcommand{\I}{{\mathcal{I}}}

\newtheorem{theorem}{Theorem}[section]
\newtheorem{corollary}{Corollary}[section]
\newtheorem{proposition}{Proposition}[section]
\newtheorem{lemma}{Lemma}[section]
\newtheorem{sublemma}{Sublemma}[section]
\theoremstyle{definition}
\newtheorem{definition}{Definition}[section]

\newtheorem{remark}{Remark}[section]

\newcommand{\eps}{\varepsilon}

\newcommand\subscr[2]{#1_{\textup{#2}}}

\newcommand{\longthmtitle}[1]{\mbox{}\textup{\textbf{(#1):}}}

\newcommand{\argmax}{\mathrm{argmax}}
\newcommand{\N}{\mathcal{N}}

\newcommand{\innerp}[2]{ \langle #1, #2\rangle}

\newcommand{\oprocendsymbol}{\hbox{$\diamond$}}
\newcommand{\oprocend}{\relax\ifmmode\else\unskip\hfill\fi\oprocendsymbol}

\newcommand{\Id}{\mathrm{Id}}

\newcommand{\Conv}{\mathrm{Conv}}
\newcommand{\dmax}{\subscr{d}{max}}
\newcommand{\dmin}{\subscr{d}{min}}

\newcommand{\A}{\mathbf{A}}
\newcommand{\HK}{\subscr{\mathbf{A}}{HK}}
\newcommand{\diag}{\mathrm{diag}}

\newcommand{\dist}{\mathrm{dist}}

\begin{document}

\begin{abstract}
We introduce a new discrete-time attention model, termed the localmax dynamics, which interpolates between the classic softmax dynamics and the hardmax dynamics, where only the tokens that maximize the influence toward a given token have a positive weight. 
As in hardmax, uniform weights are determined by a parameter controlling neighbor influence, but the key extension lies in relaxing neighborhood interactions through an alignment-sensitivity parameter, which allows controlled deviations from pure hardmax behavior. As we prove, while the convex hull of the token states still converges to a convex polytope, its structure can no longer be fully described by a maximal alignment set, prompting the introduction of quiescent sets to capture the invariant behavior of tokens near vertices. We show that these sets play a key role in understanding the asymptotic behavior of the system, even under time-varying alignment sensitivity parameters. We further show that localmax dynamics does not exhibit finite-time convergence and provide results for vanishing, nonzero, time-varying alignment-sensitivity parameters, recovering the limiting behavior of hardmax as a by-product. Finally, we adapt Lyapunov-based methods from classical opinion dynamics, highlighting their limitations in the asymmetric setting of localmax interactions and outlining directions for future research.
\end{abstract}

\title{Localmax dynamics for attention in Transformers and its asymptotic behavior}

\author[Henri Cimeti\`{e}re]{Henri Cimeti\`{e}re}
\address{Mathematical and Computer Engineering, Ecole Nationale des Ponts et Chaussées}
\email{henri.cimetiere@eleves.enpc.fr}
\author[Maria Teresa Chiri]{Maria Teresa Chiri}
\address{Department of Mathematics and Statistics, Queen's University}
\email{maria.chiri@queensu.ca}
\author[Bahman Gharesifard]{Bahman Gharesifard}
\address{Department of Mathematics and Statistics, Queen's University}
\email{bahman.gharesifard@queensu.edu}

\maketitle


\section{Introduction}
The principle of attention was first introduced by Vaswani et al. in 2017 with the publication of their seminal work~\cite{AV-NS-NP-JU-LJ-AG-LK-IP:17}, which laid the foundation for the Transformer architecture. Transformers have enabled significant progress on tasks that were previously challenging for older neural network architectures. Today, they are widely applied across diverse domains, including natural language processing (e.g., ChatGPT, Google Translate, Grammarly), computer vision (e.g., object classification and detection), multilingual speech recognition, and many other areas. Despite their widespread use, our understanding of the internal structure and behavior of trained Transformers lags behind their practical successes. As an initial step toward bridging this gap,~\cite{CY-SB-ASR-SJR-SK:19} establishes the expressive power of Transformer models by proving that they can universally approximate any continuous sequence-to-sequence function on a compact domain. In recent years, a growing body of work has sought to develop mathematical models that describe the dynamics of attention mechanisms. This line of inquiry was initiated in~\cite{YL-ZL-DH-ZS-BD-TQ-LW-TYL:19}, which proposed the first interacting particle system to model Transformer dynamics. A simplified version of this model was later introduced in~\cite{BG-CL-YP-PR:23}, where the authors derived a continuous-time attention model based on the softmax function. The resulting dynamics are given by: 
\begin{equation}
\begin{aligned}
    \dot{x}_i(t)=\sum_{j = 1}^n \frac{e^{\langle Q(t)x_i(t), K(t)x_j(t) \rangle}}{\sum_{k = 1}^n e^{\langle Q(t)x_i(t), K(t)x_k(t) \rangle}} V(t)x_j(t) = \sum_{j=1}^n P_{ij}(t) V(t)x_j(t)
\end{aligned}
\label{eq:softmax}
\end{equation}
where $x_i(t)\in \mathbb{R}^d$ refers to the $i$-th token\footnote{We interchangeably use the wordings ``token'' and ``particle'' throughout the paper.} and the matrices $Q(t)$, $K(t)$ and $V(t)$ are, respectively, the query, the key and the value matrices at a given time $ t \geq 0 $. The matrix $P\in \mathbb{R}^{n \times n }$ with general term $P_{ij}(t)$ is called the self-attention matrix. We will refer to this dynamic as the \textit{ softmax} dynamics, since the weights in the derivative of the $i$-th particle are given by the softmax function applied to the vector $\left(\langle Q x_i, Kx_1 \rangle , \cdots, \langle Qx_i, Kx_n \rangle \right)$. In~\cite{MS-PA-MB-GP:22}, the self-attention matrix is normalized using Sinkhorn’s
algorithm instead of the softmax operator in order to enforce a priori the doubly stochastic structure (i.e., a matrix whose rows and columns both sum to one) aspect on $P$. This is motivated by the fact that the self-attention matrix in the softmax dynamics have been observed converge to a doubly stochastic matrix during the training process. On a similar note, it is observed in~\cite{WS-LBZ-KM-FH-MH:20} that the trained self-attention matrix is close to a low rank matrix, motivated their Transformers model called \textit{Linformers} re-designing the attention core to enforce a priori the low rank structure, leading the decreasing the global complexity of attention from $O(n^2)$ to $O(n)$. In the same idea of dealing with the quadratic complexity of attention,~\cite{AV-AK-FF:20} developed the \textit{clustered attention} which consist in grouping ``queries into clusters and computes attention just for the centroids''. This method somehow anticipates the clustering phenomenon that naturally occurs during the attention process, as we will study in this paper. More generally, a comprehensive classification of the different variants of Transformers is established in~\cite{TL-YW-XL-XQ:21}. Finally, it is worth pointing out that without skip connection, the self attention matrix converge with a doubly exponential rate to a rank one matrix, demonstrating the issue of rank collapse~\cite{YD-JBC-AL:23}.

Going back to~\eqref{eq:softmax}, following the literature, we will assume that the matrices $Q$, $K$ and $V$ are independent from the layer $t$. This is justified in~\cite{ZL-MC-SG-KG-PS-RS:20}, where it is shown that the use of parameter-sharing across layers significantly reduce the number of parameters without seriously hurting performance. By studying the corresponding continuity equation, the authors in~\cite{BG-CL-YP-PR:23} explain how the dynamics between particles can be seen as a flow map on measure spaces. The asymptotic behavior and the emergence of clusters for dynamics~\eqref{eq:softmax} has been studied by the same authors in~\cite{BG-CL-YP-PR:24,ARA-JPS-PT:24} with various assumption on the matrices $Q$, $K$ and $V$. In~\cite{BG-CL-YP-PR:24}, there is no layer normalization, however, the result presented is on a \textit{rescaled tokens} given by $z_i(t) = e^{-V t}x_i(t)$. In~\cite{ARA-JPS-PT:24}, the layer normalization is made by projecting each token to the unit sphere, for some norm to be defined. In particular, for $V = \Id$ and $Q^TK$ 
symmetric and positive defined, it has been shown in~\cite{BG-CL-YP-PR:24} that for any initial distribution, the rescaled tokens converge toward some points on the boundary of some convex polytope $K \subset \mathbb{R}^d$. 

Closely related to our work here, a simpler discrete-time dynamics is considered in~\cite{AA-GF-EZ:24} with the constant value matrix $V = \alpha\Id$, where $ \Id$ is the identity matrix, for some parameter $\alpha > 0$. The layer normalization is made by dividing by a constant factor given by $1+\alpha$. The dynamics is given by
\begin{align}
       x_i(t+1) &= \frac{1}{1+\alpha} \left( x_i(t) + \alpha \frac{1}{|C_i(t)|}\sum_{j \in C_i(t)} x_j(t) \right) \notag\\&= x_i(t) + \frac{\alpha}{1+\alpha} \frac{1}{|C_i(t)|}\sum_{j \in C_i(t)} \left(x_j(t) - x_i(t) \right).
    \label{eq:pure_hardmax}
\end{align}

Here,
\begin{equation}
    C_i(t) = \{ j \in [n] \; | \; \langle Ax_i(t), x_j(t) \rangle = \max_{k \in [n]} \langle A x_i(t), x_k(t) \rangle \}, 
    \label{eq:neigh-hardmax}
\end{equation}
where $A$ is a symmetric positive defined matrix parameterizing the transformer model, and is taken to be $A = Q^TK$. To describe the intuition behind this model, note that the quantity of interest to determine if $x_j$ influences $x_i$ is $\langle Ax_i, x_j\rangle$, meaning that the greater this quantity is, the greater the influence of $x_j$ on $x_i$ is. The main difference between the two dynamics (apart from the discreet-time) is that in~\eqref{eq:softmax}, the particles $x_j$ with low influence toward $x_i$ have a small positive weight whereas in~\eqref{eq:pure_hardmax}, only the particles that maximize the influence toward $x_i$ have a positive weight; we will refer to this second dynamics as \textit{hardmax dynamics.} In that sense,~\eqref{eq:softmax} can be seen as a regularization of~\eqref{eq:pure_hardmax}. The asymptotic behavior of particles following (2) has been studied in \cite{AA-GF-EZ:24}. In particular, it has been shown that there exists a convex polytope $K \subset \mathbb{R}^d$ such that any $x_i$ converges to a vertex of $K$ or the projection of the origin onto some facet of $K$. The notion of \textit{leaders} (particles $x_i$ such that $C_i(t) = \{i\}$ for some $t\ge 0$) is also introduced. This is a similar result to the one about dynamics~\eqref{eq:softmax}, but the simplicity of the hardmax model allows for a more precise characterization of the structure of the limit points, see~\cite[Theorem~1.1]{AA-GF-EZ:24}.  

Related to our work below, and as already pointed out in~\cite{YL-ZL-DH-ZS-BD-TQ-LW-TYL:19}, the attention models introduced above for Transformers are closely related to some opinion dynamics models, in particular,  the Hegselmann-Krause (HK) dynamics~\cite{UK:00, RH-KU:02}. The clustering mechanism for this model has been studied in many references, including~\cite{PEJ-MS:14, VB-JH-JT:07, AB-MB-BC-HLN:13, AN-BT:12-cdc}. More specifically, under appropriate conditions, convergence to an opinion profile consisting of consensus subgroups has been established in~\cite{JL:05, LM:05}. The fact that the HK dynamics terminate in finite time was shown in~\cite{VDB-JMH-JNT:09}. Additionally, upper bounds on the termination time have been derived in both the one-dimensional and multi-dimensional settings in~\cite{AN-BT:2011, SRE-TB-AN-BT:2013}, respectively. 

\subsection*{Statement of contributions}
We introduce a new discrete-time model, termed as \emph{localmax dynamics}, which comparing to~\eqref{eq:pure_hardmax} better mimics the
softmax dynamics~\eqref{eq:softmax}. The key idea behind this model is to relax the notion of neighborhood~\eqref{eq:neigh-hardmax} by introducing an alignment sensitivity parameter, which can be designed and allows for deviations from the hard constraints in~\eqref{eq:pure_hardmax} that only the particles that maximize the influence toward a token have a positive weights. Similar to what is known for~\eqref{eq:pure_hardmax}, we prove that under this new dynamics, the convex hull of the tokens' states converges to a convex polytope. We define the notion of maximal alignment set as the state of points in this convex polytope aligned maximally with a vertex. After providing a precise characterization of this set, we show that for the localmax dynamics -- in contrast to the hardmax and softmax dynamics -- the structure of this convex polytope cannot be solely characterized by the maximal alignment set. Given this, and to gain insight into the behavior of the trajectories of the tokens in regions near the vertices of, we define the notion of 
quiescent set and show that tokens in this set will become invariant with time. As an important by-product, we prove that the localmax dynamics we have introduced will never have finite time convergence. We also prove the convergence to the maximal alignement set in the regime where the alignment sensitivity parameter is time-varying, and approaches zero. Besides the fact that this recovers the known results on convergence of the hardmax dynamics to maximal alignment set, the result reveals non-trivial facts about the localmax dynamics with time-varying alignment sensitivity parameter. As our final contribution, we examine the Lyapunov analysis commonly employed in the study of Hegselmann-Krause dynamics, now in the context of localmax dynamics. In this setting, the Lyapunov function leverages the structure of the adjoint dynamics. While this approach yields useful insights into the behavior of the adjoint vector relative to neighborhood clusters, the inherent asymmetry in the localmax neighborhoods poses significant challenges to further progress using this method. We conclude by discussing these difficulties and outlining several open directions for future research.

\section{New dynamics and problem statement}

In this paper, we aim to investigate the asymptotic behavior of the following discrete-time dynamics:

\begin{equation}
    x_i(t+1) = x_i(t) + \frac{\alpha}{1+\alpha} \frac{1}{|C_j^\delta(t)|} \sum_{j \in C_i^\delta(t)} \left( x_j(t) - x_i(t) \right) 
    \label{eq:hardmax_delta}
\end{equation}
where $\alpha$ and $\delta$ are fixed positive parameters of the model,
\begin{equation}
    C_i^\delta(t) = \{ j \in [n] \; | \; \max_{k \in [n]} \langle Ax_i(t), x_k(t) \rangle - \langle A x_i(t), x_j(t) \rangle \leq \delta \lVert Ax_i(t) \rVert \}
    \label{eq:neigh-localmax}
\end{equation}
is the neighborhood of $x_i$ at time $t \in \mathbb{N}$, and $A \in \real^{d \times d}$ is a given symmetric positive matrix. We sometimes refer to the $ \delta $ as the \emph{alignment sensitivity parameter}. As in \cite{BG-CL-YP-PR:24}, we suppose $A$ to be time-independent, and will be taken to be $A = Q^TK$ for most parts of what follows. We begin by outlining the intuition behind this model and explaining its significance.
\subsection{Motivations}
First, note that the dynamics followed by the rescaled tokens in~\eqref{eq:softmax} with $V = \Id$ and $A = Q^TK$ is given by:
\begin{equation}
    \dot{z}_i(t) = \sum_{j =1}^n \frac{e^{e^{2t}\langle Az_i(t), z_j(t) \rangle}}{\sum_{k = 1}^n e^{e^{2t}\langle Az_i(t), z_k(t) \rangle}} \left( z_j(t) - z_i(t) \right). 
    \label{eq:softmax_rescaled}
\end{equation}
We observe that a scaling factor $e^{2t}$ appears in front of the quantity $\langle A z_i(t), z_j(t)\rangle$. As a result, for $t$ large, the only non-negligible weights are associated with particles $x_j$ maximizing $\langle A z_i(t), z_j(t)\rangle$. This is similar to what happens in dynamics~\eqref{eq:pure_hardmax} but, at the beginning (for $t$ small), dynamics~\eqref{eq:softmax} is richer in the sense that it allows a wider range of movement for the particles. For instance, a particle satisfying $\mbox{argmax}_{k \in [n]} \langle Ax_i(t), x_k(t) \rangle = \{ i \}$ will be stationary for dynamics~\eqref{eq:pure_hardmax}, in contrast to~\eqref{eq:softmax}. 

\begin{figure}[h]
\centering
\includegraphics[width=0.8\textwidth]{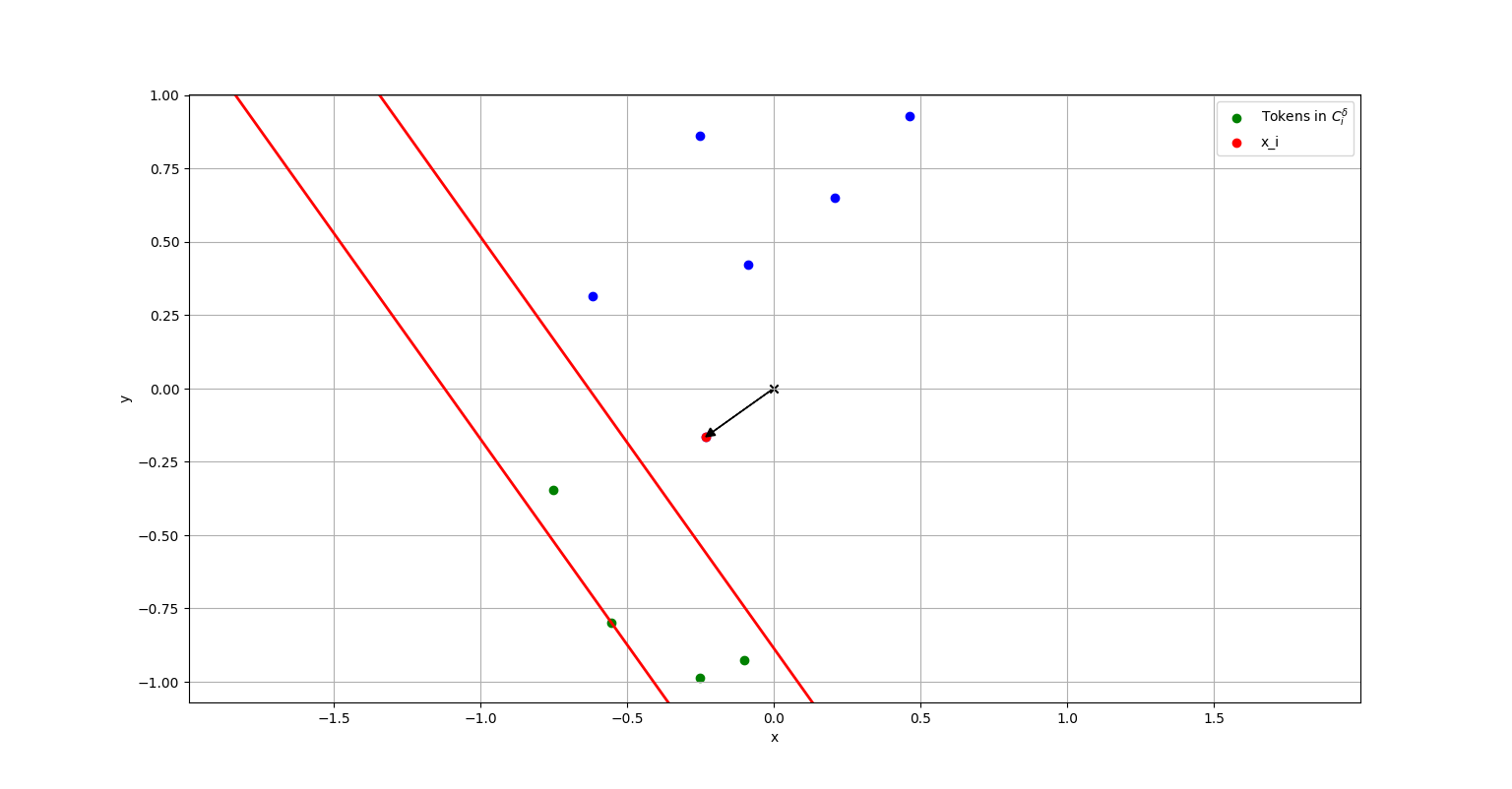}
\captionsetup{justification=centering}
\caption{Representation of the set $C_i^\delta$ (in green) in the case $d=2$ and $A = I_2$. The two red lines correspond to the border of $Z_i$.}
\label{fig:1}
\end{figure}

Clearly, limitations in capturing such behaviors impact the expressivity of the outputs generated by pure hardmax models. This motivates our model~\eqref{eq:hardmax_delta}, which replaces the neighborhood definition~\eqref{eq:neigh-hardmax} with~\eqref{eq:neigh-localmax}. We now discuss some consequences of this modification. To wit, let $j \in \mbox{argmax}_{k \in [n]} \langle Ax_i(t), x_k(t) \rangle$. Instead of selecting only the particles in $\mbox{argmax}_{k \in [n]} \langle Ax_i(t), x_k(t) \rangle$, we also select those in the strip of width $\delta$, orthogonal to $x_i(t)$ and starting from $x_j(t)$, see Figure~\ref{fig:1}. For this reason, from this point on, we refer to~\eqref{eq:hardmax_delta} as the \emph{localmax dynamics}, highlighting its contrasts with the hardmax dynamic~\eqref{eq:pure_hardmax}. It is important to note that since $\delta$ is a constant parameter, the dynamics have a different behavior asymptotically. In particular, the set of candidate convergence point identified in~\cite{BG-CL-YP-PR:23} and in~\cite{AA-GF-EZ:24} is no longer valid in our case. However, in Section~\ref{sec:delta(t)}, we show that by slightly modified our model, we retrieve the convergence in $S$. 

\begin{figure}[h]
    \centering
   
    \subfigure[Pure hardmax dynamics\label{fig:2b}]{
        \includegraphics[width=0.30\textwidth]{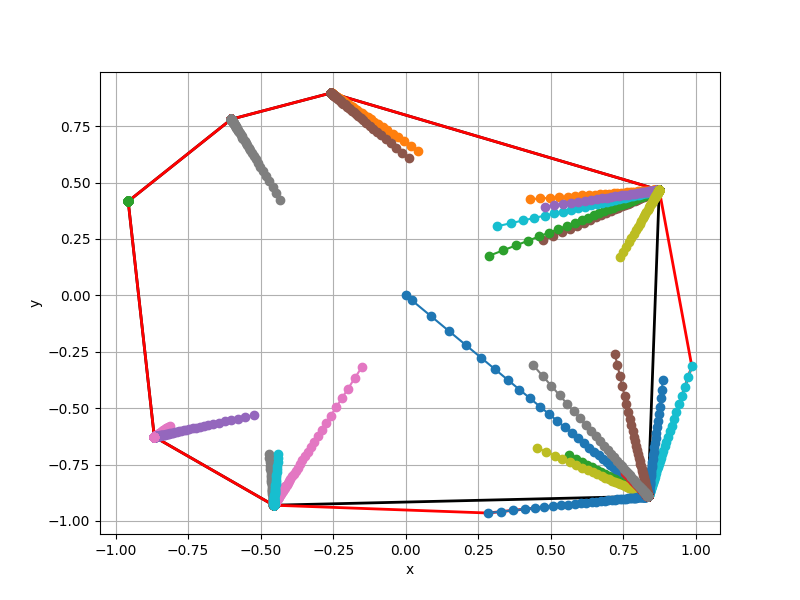}
    }    
    \subfigure[Localmax dynamics with $\delta = 0.3$\label{fig:2c}]{
        \includegraphics[width=0.30\textwidth]{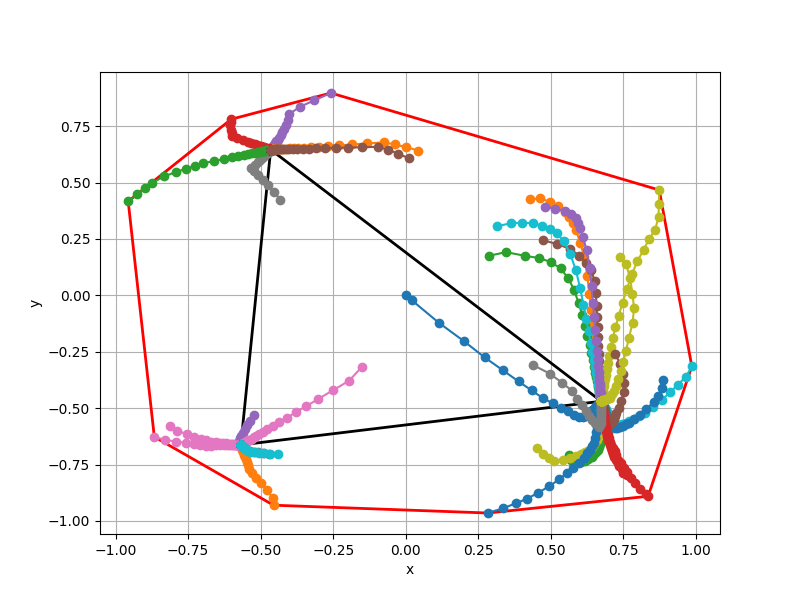}
    }
    \subfigure[Softmax dynamics\label{fig:2a}]{
        \includegraphics[width=0.30\textwidth]{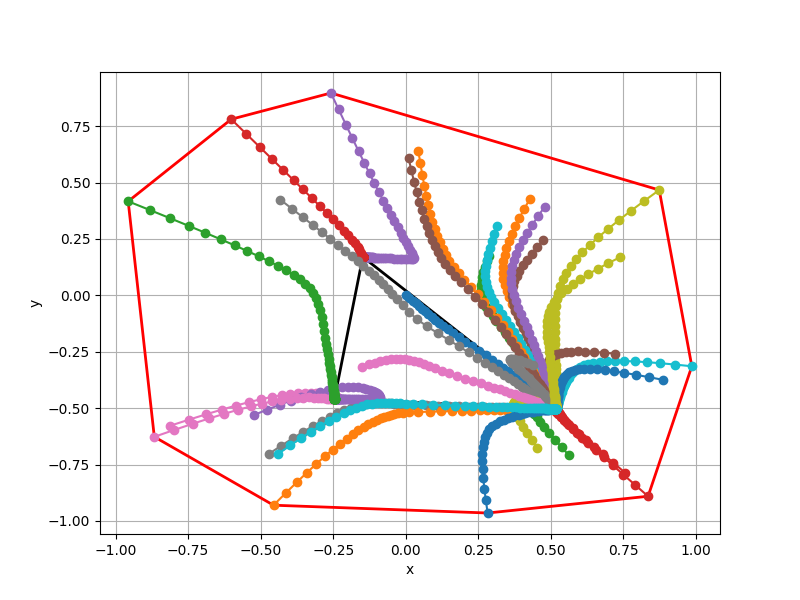}
    }     
    \captionsetup{justification=centering}
    \caption{Comparison between the trajectories for the different dynamics for the same random initial distribution ($n = 30$ and $d=2$). The convex hull of the initial (resp. final) distribution is represented in red (resp. black).}
    \label{fig:2}
\end{figure}

It is convenient to denote by 
    \begin{equation}
        Z_i(t) := \{ x \in \mathbb{R}^d \; | \; \max_{k \in [n]} \langle Ax_i(t), x_k(t) \rangle - \langle A x_i(t), x \rangle \leq \delta \lVert Ax_i(t) \rVert \},
        \label{eq:Z_i(t)}
    \end{equation}
    such that $j \in C_i^\delta(t) \iff x_j(t) \in Z_i(t)$.
In Figure~\ref{fig:2a},~\ref{fig:2b}, and~\ref{fig:2c}, we plot the trajectories for dynamics \eqref{eq:softmax_rescaled}, \eqref{eq:pure_hardmax}, and~\eqref{eq:hardmax_delta}, respectively. We take a random initial distribution in the box $[-1,1] \times [-1, 1]$ and set the parameters $\delta = 0.3$, $\alpha = 0.2$ and $T_{max} = 100$. We also plot the initial and final convex hull of the tokens. We observe that the pure hardmax dynamics allows very little movement for the convex hull while tokens following the hardmax dynamics with strip of width $\delta$ have a behavior that is more similar to the softmax dynamics.

\begin{remark}
    \label{rem:inner_prod}
    From now on, we will always take $A = I_d$. This is because we can easily recover the general case by changing the inner product, as $A$ is always assumed to be symmetric and positive defined. \oprocend
\end{remark}

\section{Macro-level properties of the trajectories}\label{sec:macro}
In what follows, $(x_i(t))_{1 \le i \le n }$ will always refer to the position of the tokens for an arbitrary initial distribution at time $ t \geq 0$, which in turn defines the trajectories. Also, we will use the notation $\Conv(x_\ell(t), 1\le \ell \le n)$ to refer to the convex hull of the particles at time $t \in \mathbb{N}$.
\begin{theorem}\longthmtitle{Convergence of convex hull to a polytope}
    \label{thm:convex_hull_decreasing}
    The convex hull $\Conv(x_l(t), 1\leq \ell \leq n)$ converges toward a convex polytope $K \subset \mathbb{R}^d$. 
\end{theorem}
\begin{proof}
Let $t\ge 0$ and let $H$ be any closed half space of $\mathbb{R}^d$ that does not contain any $x_i(t)$. Let $\mathbf{n}$ be the unit outer normal to $H$. If $j \in \mbox{argmin}_{i \in [n]}\,\dist(x_i(t), H)$, then any particle in $C_j^\delta(t)$ is further away from $H$ than $x_j(t)$. Therefore, we have that 
$$
\langle x_\ell(t) - x_j(t), \mathbf{n}\rangle \ge 0\,,
$$
for any $\ell \in C_j^\delta(t)$. This implies that $\langle x_j(t+1) - x_j(t), \mathbf{n}\rangle \ge 0$. Thus, particles in $ \mbox{argmin}_{i \in [n]}\,\dist(x_i(t), H)$ must get further away from $H$. 
Let us now take $i \in [n] \backslash \mbox{argmin}_{\ell \in [n]}\,\dist(x_\ell(t), H)$. Since, $j \in \mbox{argmin}_{\ell \in [n]}\,\dist(x_\ell(t), H)$, we know that for any $\ell \in C_i^\delta(t)$, $\innerp{x_\ell(t) - x_i(t)}{\mathbf{n}} \ge \innerp{x_j(t)-x_i(t)}{\mathbf{n}}$. Thus, we have that
\begin{equation*}
    \begin{aligned}
        \langle x_i(t+1) - x_i(t), \mathbf{n}\rangle &= \frac{\alpha}{1+\alpha} \frac{1}{|C_i^\delta(t)|} \sum_{{\ell} \in C_i^\delta(t)} \langle x_\ell(t) - x_i(t), \mathbf{n} \rangle \\
        &\ge \frac{\alpha}{1+\alpha} \frac{1}{|C_i^\delta(t)|} \sum_{{\ell} \in C_i^\delta(t)} \langle x_j(t) - x_i(t), \mathbf{n} \rangle \\
        &= \frac{\alpha}{1+\alpha} \langle x_j(t) - x_i(t), \mathbf{n} \rangle \\
        &> - |\langle x_j(t) - x_i(t), \mathbf{n} \rangle |\,.
    \end{aligned}
\end{equation*}
This proves that 
\[
\dist(x_i(t+1), H) > \dist(x_j(t), H)\,.
\]
Therefore, the minimal distance between the particles and $H$ increases with $t$. Hence, upon rewriting 
\begin{equation*}
    \Conv(x_\ell(t), 1\leq \ell \leq n) = \bigcap_{\substack{H\ \text{closed half space} \\ 
\Conv(x_\ell(t),\, 1 \leq {\ell} \leq n) \cap H = \emptyset}} \mathbb{R}^d \backslash H\,,
\end{equation*}
it follows that $\Conv(x_\ell(t+1), 1\leq \ell \leq n) \subset \Conv(x_\ell(t), 1\leq \ell \leq n)$. This proves the claim. 
\end{proof}

From now on, we denote the limiting convex polytope in Theorem~\ref{thm:convex_hull_decreasing} by $K $, and its vertices by the set $V:=\{ v_1, ..., v_r\}$, where $ r $ is a positive integer. 
Since $K$ is a convex polytope, any element $x \in K$ can be expressed as a convex combination of the vertices of $K$ : There exist $\alpha_1, ...., \alpha_r \ge 0$ such that $\sum_{k = 1}^r \alpha_k = 1$ and $x = \sum_{k = 1}^r \alpha_k v_k$. Then, it is easy to see that $\lVert x \rVert \le \max_{k \in [r]} \langle x, v_k \rangle$.     For later use, we define 
    \begin{equation}\label{eq:eta}
    \eta(t):= \max_{k \in [n]} \dist(x_k(t), K),    
    \end{equation}    
    where $t \in \mathbb{N}$. Note that by Theorem \ref{thm:convex_hull_decreasing} implies that $\lim_{t \rightarrow +\infty} \eta(t) = 0$.

\begin{definition}\longthmtitle{Maximal alignment set}
    We define the following set of element in $K$ :
    \label{def:S_set}
    \begin{equation}
        S := \{ x \in K \; | \; \lVert x \rVert^2 = \max_{k \in [r]} \langle x, v_k \rangle \}\label{eq:maximal-alig-S}
    \end{equation}
     and refer to it as the 
    \emph{maximal alignment set}. We also define for any $x \in K$ :
    \begin{equation}
        \I_x := \{ \ell \in [r] \; | \; \innerp{x}{v_\ell} = \max_{k \in [r]} \innerp{x}{v_k} \}\,.
    \end{equation}
   
\end{definition}
As will be clear shortly, characterizing the maximal alignment set is closely related to understanding the structure of the limiting convex polytope. In fact, in the simpler dynamics considered in \cite{AA-GF-EZ:24} and \cite{BG-CL-YP-PR:24}, the trajectories of every particle eventually reaches this set. Notably, as will be discussed in more detail later, this is not the case for the localmax dynamics introduced in~\eqref{eq:hardmax_delta}. 

It is worth pointing out here that even if the particles lie in $\mathbb{R}^d$, when we use the terminology of ``border'' or ``hyperplane'', it will always be with respect to the smallest vector space $E \subset\mathbb{R}^d$ containing $K$. 

This first lemma will play a crucial role in several arguments. Roughly speaking, it asserts that after a finite time, it is impossible for all particles located in a neighborhood of some vertex 
$v \in V$ at time $t$ to leave this neighborhood simultaneously at time $t+1$. From now on, for any $\epsilon > 0$, we use the notation $K_\epsilon$ to denote the set of points in $\mathbb{R}^d$ that lie within distance at most $\epsilon$ of $K$, that is: 
\begin{equation}
    \label{eq:def_K_eps}
    K_\epsilon := \{ x \in \mathbb{R}^d \; | \; \dist(x, K) \le \epsilon \}.
\end{equation}
\begin{lemma}
    \label{lem:cant_leave_vertices_neigh}
    Let $v \in V$ and $0<\eps\le \delta$. There exists an increasing sequence of time $(t_k)_{k \in \mathbb{N}}$ going to $+\infty$ such that, for any $k \in \mathbb{N}$, there exists $i \in [n]$ satisfying $x_i(t_k) \in B(v, \eps)$ and $x_i(t_k + 1) \in B(v, \eps)$
\end{lemma}
\begin{proof}
    Let $Q:= \{ x \in \mathbb{R}^d \; | \; \innerp{x-v}{u} = 0 \}$ where $u$ is a unit vector such that $Q$ supports $K$ in $v$ and in $v$ only, i.e., 
            $K \subseteq \{x \in \mathbb{R}^d : \langle x-v,u\rangle \ge 0\} $ and 
$K \cap Q = \{v\}$. There exists a constant $\mu > 0$ such that if $x \in K_\mu$, where $ K_\mu $ is defined as in~\eqref{eq:def_K_eps}, satisfies $\innerp{x-v}{u}\le 0$, then $x \in B(v, \eps)$. Now, we define the quantity 

\begin{equation}\label{eq:rhomu}
    \rho_\mu:= \inf_{y \in K_\mu \backslash B(v, \eps)} \innerp{y-v}{u}\ge 0.
\end{equation}
Upon replacing $\mu$ by $\frac{\mu}{2}$, and using compactness of $K_\mu$ and the supporting property of $Q$,
we may assume that $\rho_\mu>0$. 
            We now define $T_1\ge 0$ such that $\eta(t) \le \mu$ for any $t \ge T_1$. Because of the decrease of the convex hull for the inclusion, there exists at least one particle $x_i$ satisfying $\innerp{x_i(t) - v}{u} \le 0$ for any given time $t\ge0$. This is equivalent to say that 
            \begin{equation}\label{eq:mt}
            m(t):= \min_{\ell \in [n]} \innerp{x_\ell(t)-v}{u} \le 0.
            \end{equation}
            Also, the convergence of the convex hull to $K$ gives $m(t) \to 0$ as $t\to \infty$.\\ \\
            We are now in position to conclude the proof. Suppose by contradiction that after finite time $T_2 \ge T_1$, all particles in $B(v, \eps)$ at time $t \ge T_2$ will leave this set at time $t+1$. Let $t \ge T_2$ and take $i \in [n]$ such that $\innerp{x_i(t+1)-v}{u}\le 0$. Since, $t\ge T_1$, we know that $x_i(t+1) \in B(v, \eps)$ and since $t \ge T_2$, we know that $x_i(t)\in K_\mu \backslash B(v, \eps)$. Using~\eqref{eq:mt}, we have that
            \begin{equation*}
            \begin{aligned}
                    \innerp{x_i(t+1) - &x_i(t)}{u} = \frac{\alpha}{1+\alpha} \frac{1}{|C_i^\delta(t)|} \sum_{j \in C_i^\delta(t)} \innerp{x_j(t)-x_i(t)}{u} \\
                    &\ge \frac{\alpha}{1+\alpha} \min_{\ell \in [n]} \innerp{x_{\ell}(t)-x_i(t)}{u}
                    =\frac{\alpha}{1+\alpha}m(t) + \frac{\alpha}{1+\alpha}\innerp{v-x_i(t)}{u}
            \end{aligned}
            \end{equation*}
            Therefore:
            \begin{align}
                \innerp{&x_i(t+1) - v}{u} = \innerp{x_i(t+1) - v}{u} -\frac{1}{1+\alpha} \rho_\mu + \frac{1}{1+\alpha} \rho_\mu \\
                &\ge \innerp{x_i(t+1) - v}{u} - \frac{1}{1+\alpha} \innerp{v - x_i(t)}{u} + \frac{1}{1+\alpha} \rho_\mu
                \ge \frac{\alpha}{1+\alpha}m(t) + \frac{1}{1+\alpha} \rho_\mu
                \label{eq:aux32}
            \end{align}
            It only remains to define $T_3 \ge T_2$ such that $|m(t)| < \frac{\rho_\mu}{\alpha}$ for any $t\ge T_3$. Thus, for $t\ge T_3$, if $x_i(t) \notin B(v, \eps)$, then~\eqref{eq:aux32} simplifies to $\innerp{x_i(t+1)-v}{u} > 0$.

            To summarize what we just proved: for $t \ge T_3$, we know by assumption that all the particles in $B(v, \eps)$ at time~$t$ will leave this set at time $t+1$. Moreover, the particles outside $B(v, \eps)$ at time $t$ cannot satisfy $\innerp{x_i(t+1)-v}{u} \le 0$.
            This proves that 
             no particle \emph{at all} can satisfy $\innerp{x_i(t+1)-v}{u} \le 0$ at time $t+1$, which is a contradiction, because by definition~\eqref{eq:mt}, $m(t) \le 0$ for all $t$ meaning that            there always exists at least one index $i$ for which 
\[
\langle x_i(t) - v, u \rangle \le 0.
\]             
\end{proof}
We still need a few intermediary results. Lemma~\ref{lem:no_influence_between_vertices}, combined with Lemma~\ref{lem:cant_leave_vertices_neigh}, states that two particles located near two different vertex of $K$ cannot influence each other. 
\begin{lemma}
    \label{lem:no_influence_between_vertices}
    There exists a time $T$ such that, for any $t\ge T$ and any $v,v' \in V$ with $v \ne v'$, if $x_i(t) \in B(v, \eta(t))$, $x_j(t) \in B(v', \eta(t))$ and $j \in C_i^\delta(t)$, then $x_i(t+1) \notin B(v, \eta(t))$.
\end{lemma}
\begin{proof}
            Suppose we can find $x_i(t) \in B(v, \eta(t))$ and $x_j(t) \in B(v', \eta(t))$ satisfying $j \in C_i^\delta(t)$. 
            Note that for the vertices of the polytope, 
            we can find $r$ unit vectors $(n_k)$ such that for any $1\le k,\ell \le r$, $k\ne l$:
            \begin{equation}
                \langle v_k, n_k \rangle > \innerp{v_\ell}{n_k}\,.
            \end{equation}
            Let $v \in V$ and let $n$ be the unit vector associated with $v$. Denote $S = \innerp{v}{n}$, $s = \max_{k \in [r], v_k \ne v } \innerp{v_k}{n}$ and define $c = \frac{s + S}{2}$. Let $H = \{ x \in \mathbb{R}^d \; | \; \innerp{x}{n} = c \}$. We know that 
            $$\dist(v, H) = \min_{k \in [r], v_k \ne v } \dist(v_k, H) = \frac{S-s}{2}:= D.$$
            Then, the two following statements hold:
            \begin{itemize}
                \item[-] $\innerp{x_\ell(t)-x_i(t)}{n} \le 2 \eta(t)$ for any $\ell \in [n]$,
                \item[-] $\innerp{x_j(t)-x_i(t)}{n} < -2(D - \eta(t))$ as long as $\eta(t) \le D$.
            \end{itemize}
            This is because all particles are at a distance of at most $\eta(t)$ from $K$, with $x_i(t) \in B(v,\eta(t))$ and $x_j(t) \in B(v', \eta(t))$. Therefore,
            \begin{equation}
            \label{eq:bound_inner_prod}
            \begin{aligned}
                \innerp{x_i(t+1) - x_i(t)}{\mathbf{n}} &= \frac{\alpha}{1+\alpha} \frac{1}{|C_i^\delta(t)|} \sum_{k \in C_i^\delta(t)} \innerp{x_k(t) - x_i(t)}{\mathbf{n}} \\
                &\le \frac{\alpha}{1+\alpha} \frac{1}{|C_i^\delta(t)|} \left( 2\eta(t) - 2D + 2 (|C_i^\delta(t)| - 1) \eta(t) \right) \\
                &\le \frac{2\alpha}{1+\alpha} \frac{1}{|C_i^\delta(t)|} \left(|C_i^\delta(t)| \eta(t) -D\right)
                \le \frac{2\alpha}{1+\alpha} \left(\eta(t) - \frac{1}{n}D\right). 
            \end{aligned}
            \end{equation}
            Consequently, we can define $T$ such that $\eta(t) \le \frac{\alpha}{1+2\alpha}\frac{D}{n}$ for any $t \ge T$ and state that if the previous setup holds for $t \ge T$, then $x_i(t+1) \notin B(v, \eta(t))$. We then conclude using Lemma~\ref{lem:cant_leave_vertices_neigh}.
\end{proof}
\begin{theorem}\longthmtitle{Characterization of maximal alignment set}
    \label{thm:S_carac}
     The following properties hold true :
     \begin{enumerate}
         \item The set $S$ is finite.
         \item $S \subset \partial K \cup \{0\}$
         \item $V \subset S$
         \item The elements of $S$ that are not in $V$ are the projection of the origin to some facet of $K$ of dimension $p \ge 1$.
     \end{enumerate}
\end{theorem}
Before presenting the proof, we include some remarks to compare our results with those in~\cite{AA-GF-EZ:24} and~\cite{BG-CL-YP-PR:24}. The proof of the first statement, item (1), is new, but the ones of item (2) and item (4) follow the same lines as in~\cite[Lemma~4.1 and Lemma~4.4]{AA-GF-EZ:24}. However, crucially, the statement in item (3) is not proved in~\cite{BG-CL-YP-PR:24}, and the proof provided in~\cite{AA-GF-EZ:24} for a different dynamics uses a strong convergence result that will not hold for our dynamics. In this sense, the proof we provide here is our main contribution to the characterization of the maximal alignment set $S$.
We begin with two general results that will be of great use in the proof of Theorem~\ref{thm:S_carac}  as well as in subsequent arguments. 
\begin{lemma}
    \label{lem:max_inner_prod_near_vertices}
    Let $w \in K\backslash \{0\}$, recall that $\mathcal{I}_w:= \argmax_{k \in [r]} \innerp{w}{v_k}$ and let $F$ the face of $K$ generated by the vertices in $\mathcal{I}_w$. There exists $\eps_0 > 0$ and $c > 0$ such that for any $0< \eps \le \eps_0$ :
    \begin{enumerate}
        \item For any $x \in \{ y \in \mathbb{R}^d \; | \; \dist(y, F) \le \eps \mbox{ or } \| y - w \| \le \eps\}$, $\argmax_{k \in [r]} \innerp{x}{v_k} \subset \mathcal{I}_w$.
        \item For any $x \in B(w, \eps)$, if $y \in K_\eps$ is such that $\dist(y, F) \ge 2\eps$, then $\innerp{x}{y} \le \innerp{w}{v_j} - c$ for any $j \in \mathcal{I}_w$.
    \end{enumerate}
\end{lemma}
\begin{proof}
    Let $f$ be the function defined on $\mathbb{R}^d$ such that $$f_j(x) = \innerp{x}{v_j} - \max_{k \in [r] \backslash \mathcal{I}_w}\innerp{x}{v_k}$$ for any $j \in \mathcal{I}_w$. This function is continuous and strictly positive on $F$ but also satisfies $f(w) > 0$. Therefore, there exists $\eps_1 > 0$ such that for any $0<\eps\le\eps_1$, $f(x) > 0$ for any $x \in \mathbb{R}^d$ such that $\dist(x, F) \le \eps$ and for any $x \in B(w, \eps)$. This means that for any $x$ close enough to $F \cup \{w \}$, the vertices maximizing the scalar product with $x$ are in $\mathcal{I}_w$. \\
    Now, we know that : 
    \begin{equation}
    \label{eq:lemma3.2_eq_1}
        \left( y \in K_\eps \; \; \mathrm{and } \; \; \dist(y, F) \ge 2\eps \right) \implies \left( \forall y_0 \in K, \; \| y - y_0 \|\le \eps \implies \dist(y_0, F) \ge \eps\right)\,.
    \end{equation}
    This is because $2\eps \le \dist(y, F) \le \dist(y_0, F) + \| y - y_0\| \le \dist(y_0, F) + \eps$. Furthermore, we can define the mapping $ g : \mathbb{R}^d \rightarrow \mathbb{R}$ by
    \begin{equation*}
         x \mapsto   \sup_{z \in K, \dist(z, F)\ge \eps} \innerp{x}{z}.
    \end{equation*}
    Note that $g$ is continuous and satisfies $g(w) \le \innerp{w}{v_j} - c_0$ for some $c_0 > 0$ and for any $j \in \mathcal{I}_w$. This is because $\innerp{w}{\cdot}$ is continuous, the set $\{ z \in K, \; \dist(z, F) \ge \eps\}$ is compact, and $\innerp{w}{\cdot}$ is strictly less than $\innerp{w}{v_j}$ on this set. Therefore, since any closed ball around $w$ is compact, by continuity, we can define $\eps_2$ such that :
    \begin{equation}
        \label{eq:lemma3.2_eq_2}
        g(x) \le \innerp{w}{v_j} - \frac{c_0}{2},\;\; \mbox{for any}\; \; x \in B(w, \eps_2)\,.
    \end{equation}    
    We are now in position to conclude the proof. Define $\eps_3$ such that 
    \[
    \eps \le \frac{c_0}{2(\max_{s \in K}\|s\|+\eps)},
    \]
    for any $\eps \le \eps_3$. 
    Let $\eps_0 := \min(\eps_1, \eps_2, \eps_3)$ and let $0 < \eps \le \eps_0$. If $x \in B(w, \eps)$:
    \begin{itemize}
        \item Since $\eps \le \eps_1$, we know that $\argmax_{k \in [r]} \innerp{x}{v_k} \subset \mathcal{I}_w$.
        \item Suppose $y \in K_\eps$ and $\dist(y, F) \ge 2\eps$. There exists $y_0 \in K$ such that $\| y - y_0 \| \le \eps$. Using~\eqref{eq:lemma3.2_eq_1}, we know that $\dist(y_0, F) \ge \eps$. Therefore, using~\eqref{eq:lemma3.2_eq_2}, if $j \in \mathcal{I}_w$, it follows $\innerp{x}{y_0} \le \innerp{w}{v_j} - \frac{c_0}{2}$. Then :
        \begin{equation*}
            \begin{aligned}
                \innerp{x}{y} &= \innerp{x}{y_0} + \innerp{x}{y-y_0} \\
                &\le \innerp{w}{v_j} - \frac{c_0}{2} + (\|w\|+\eps)\eps
                \le \innerp{w}{v_j} - c
            \end{aligned}
        \end{equation*}
        where $\frac{c_0}{2} - (\|w\|+\eps) \eps \ge \frac{c_0}{2} - (\max_{s \in K}\|s \| + \eps_3) \eps_3 := c  > 0$ because $\eps \le \eps_3$.
    \end{itemize}
This concludes the proof.     
\end{proof}
\begin{lemma}
\label{lem:diff_max}
    For any $t \in \mathbb{N}$ and $i \in [n]$, $$0 \leq \max_{\ell \in [n]} \langle x_i(t), x_\ell(t) \rangle - \max_{\ell \in [r]}\langle x_i(t), v_\ell \rangle \leq \eta(t)\lVert x_i(t) \rVert\,,$$ where $ \eta(t) $ is as in~\eqref{eq:eta}. 
\end{lemma}
\begin{proof}

    The left inequality is obvious since $K \subset \Conv(x_\ell, \ell \in [n])$. Now, take $j$ such that $\max_{\ell \in [n]} \langle x_i(t), x_\ell(t) \rangle = \langle x_i(t), x_j(t) \rangle$ and suppose that $\langle x_i(t), x_j(t) - v_\ell \rangle > \eta(t) \lVert x_i(t) \rVert$ for any $1 \leq \ell \leq m$. For any $z = \sum_{\ell \in [r]} \alpha_\ell v_\ell \in K$, we have :
\begin{equation*}
   \lVert x_i(t) \rVert \lVert x_j(t) - z \rVert \ge \langle x_i(t), x_j(t) - z \rangle = \sum_{\ell \in [r]} \alpha_\ell \langle x_i(t), x_j(t) - v_\ell \rangle > \eta(t) \lVert x_i(t) \rVert
\end{equation*}
which proves $\dist(x_j(t), K) > \eta(t)$ which is not true by definition. This concludes the proof.
\end{proof}
We are now in a position to provide the proof of Theorem~\ref{thm:S_carac}.

\begin{proof}[Proof of Theorem~\ref{thm:S_carac}] We proceed with the proof of each item: 
    \begin{itemize}    
            \item[(1)] $\I_x$ is a subset of $[r]$. Thus, if we prove that 
            \[
            \I_x = \I_y \implies x=y,
            \]
            we have proved that $|S| \leq 2^m< +\infty$. To show this, let $x,y \in S$ such that $\I_x = \I_y$. Writing $x = \sum_{k \in \I_x} \alpha_k v_k$ and $y = \sum_{\ell \in \I_y} \alpha_\ell v_l$ as convex combinations, we have that 
            \begin{equation*}
                \begin{aligned}
                    \lVert x-y \rVert^2 &= \langle x-y, x-y \rangle \\
                    &= \langle x - \sum_{\ell \in \I_y} \beta_\ell v_\ell, x \rangle - \langle \sum_{k \in \I_x}\alpha_k v_k -  y , y \rangle \\
                    &= \sum_{\ell \in I_y} \beta_l\langle x - v_l, x \rangle - \sum_{k \in I_x}\alpha_k \langle v_k - y , y \rangle \\
                    &= 0\,.
                \end{aligned}
            \end{equation*}

            \item[(2)] We proceed as in \cite{AA-GF-EZ:24}. Let $x\in S$. We have 
            \[
            \langle x, x \rangle = \max_{k \in [r]}\, \langle x, v_k \rangle = \max_{y \in K} \, \langle x, y\rangle,
            \]
            which implies that $x$ maximizes the linear function $y \mapsto \langle x, y \rangle$ on the convex polytope $K$. This proves that $x \in \partial K$ or $x = 0$.

            \item[(3)] Let $v \in V$ and suppose, on the contrary to the statement, that $v\notin S$. Let $J = \argmax_{k \in [r]} \innerp{v}{v_k}$. By assumption, $v_j \ne v$ for any $j \in J$ which implies in particular that $v\ne 0$. Now, take $\eps_0$ as defined in Lemma~\ref{lem:max_inner_prod_near_vertices} with $w = v$ and take $0 < \eps \le \eps_0$. As stated by this lemma, we know that there exists $c >0$ such that for any $x \in B(v, \eps)$, if $y \in K_\eps$ is such that $\dist(y, F) \ge 2\eps$, then $\innerp{x}{y} \le \innerp{w}{v_k} - c$ for any $k \in J$.
            Since the convex hull $\Conv(x_\ell(t), \ell \in [n])$ converges towards $K$ as $ t\rightarrow \infty$, there exists a particle in the $\eta(t)-$neighborhood of any vertex for any given time $ t $. Let $T_1\ge 0$ be such that for any $t \ge T_1$, $\eta(t) \le \eps$. We fix $t \ge T_1$ and we take $i \in [n]$ such that $x_i(t) \in B(v, \eps)$ and let $j \in \argmax_{\ell \in [n]} \innerp{x_i(t)}{x_\ell(t)}$. Using Lemma~\ref{lem:diff_max}, $\innerp{x_i(t)}{x_j(t)}\ge  \max_{\ell \in [r]}\innerp{x_i(t)}{v_\ell}$. This implies:
            \begin{equation}
                \label{eq:theorem3.2item3}
                \innerp{x_i(t)}{x_j(t)} \ge \max_{\ell \in [r]}\innerp{v}{v_\ell} - \eps \max_{\ell \in [r]} \| v_\ell \|\,.
            \end{equation}
            Since $v \ne 0$, we know that $\max_{\ell \in [r]} \| v_\ell \| >0$ so, upon diminishing $\eps$ (which amount to increasing $T_1$), we can also enforce the condition 
            \begin{equation*}
                0 < \eps \le \frac{c}{2\max_{\ell \in [r]} \| v_\ell \|}
            \end{equation*}
            which, using~\eqref{eq:theorem3.2item3}, finally gives
            \begin{equation*}
                \innerp{x_i(t)}{x_j(t)} > \max_{\ell \in [r]}\innerp{v}{v_\ell} - c
            \end{equation*}
            and item (2) of Lemma~\ref{lem:max_inner_prod_near_vertices} gives that $\dist(x_j(t), F) \le 2\eps$.
            By assumption, $v \notin F$. In particular, $\| p_F(v) - v \| > 0$ where $p_F$ is the orthogonal projection to $F$. Let $\mathbf{n} = \frac{p_F(v) - v}{\| p_F(v) - v \|}$ and let $j \in \argmax_{\ell \in [n]} \innerp{x_i(t)}{x_\ell(t)}$. We write :
            \begin{equation*}
            \begin{aligned}
                \innerp{p_F(v) - x_j(t)}{\mathbf{n}} &=  \innerp{p_F(v) - p_F(x_j(t)) + p_F(x_j(t)) - x_j(t)}{\mathbf{n}} \\
                &=\innerp{p_F(x_j(t)) -x_j(t)}{\mathbf{n}} \\
                &\le \|p_F(x_j(t)) - x_j(t) \| \\
                &= \dist(x_j(t), F) \\
                &\le 2\eps
                \end{aligned}
            \end{equation*}
            and then:
            \begin{equation*}
                \begin{aligned}
                    \innerp{x_j(t)-x_i(t)}{\mathbf{n}} &= \innerp{p_F(v) - v - (p_F(v) - x_j(t)) + (v - x_i(t))}{\mathbf{n}} \\
                    &= \innerp{p_F(v) - v}{\mathbf{n}} - \innerp{p_F(v) - x_j(t)}{\mathbf{n}} + \innerp{v - x_i(t)}{\mathbf{n}} \\
                    &\ge \innerp{p_F(v) - v}{\mathbf{n}} - 2\eps - \eps \\
                    &= \| p_F(v) - v \| - 3\eps
                \end{aligned}
            \end{equation*}
             and also :
            \begin{equation*}
                \begin{aligned}
                    \| x_i(t+&1) - x_i(t) \|\\[2ex]& \ge \innerp{x_i(t+1) - x_i(t)}{\mathbf{n}} \\
                    &= \frac{\alpha}{1+\alpha} \frac{1}{|C_i^\delta(t)|} \sum_{\ell \in C_i^\delta(t)} \innerp{x_\ell(t)-x_i(t)}{\mathbf{n}}\\
                    &= \frac{\alpha}{1+\alpha} \frac{1}{|C_i^\delta(t)|} \left( \innerp{x_j(t) - x_i(t) }{\mathbf{n}} + \sum_{\ell \in C_i^\delta(t) \backslash \{j \}} \innerp{x_\ell(t) - x_i(t)}{\mathbf{n}} \right)\,. \\
                \end{aligned}
            \end{equation*}
            Now, note that, since $v$ is a vertex of $K$, $H:= \{x \in \mathbb{R}^d \; |\; \innerp{x-v}{\mathbf{n}} \le 0\}$ is a half space supporting $K$ on $v$. Therefore, if $y \in K_\eps$, then $\innerp{y-v}{\mathbf{n}}\ge -\eps$, which in turn implies that $\innerp{x_\ell(t)-x_i(t)}{\mathbf{n}} \ge -2 \eps$ for any $\ell \in [n]$. Combining the two preceding inequalities, we conclude that 
            \begin{equation}
            \begin{aligned}
                \| x_i(t+1) - x_i(t) \| &\ge \frac{\alpha}{1+\alpha} \frac{1}{|C_i^\delta(t)|} \left( \| p_F(v) - v \| - 3\eps - 2(|C_i^\delta(t)|-1) \eps\right) \\
                &= \frac{\alpha}{1+\alpha} \left( \frac{\| p_F(v) - v \|}{n} - 3 \eps \right)\,.
                \end{aligned}
            \end{equation}
            Recall that this is true as long as $\eps \le \eps_0$ and $\eta(t) \le \eps$ i.e., $t \ge T_1$. Thus, we can take $0< \eps \le \eps_0$ such that 
            \begin{equation*}
                \eps < \frac{\|v - p_F(v)\|}{n} \frac{\alpha}{5\alpha + 2}.
            \end{equation*}
            Note that $T_1$ depends on $\eps$ so diminishing $\eps$ amount to increasing $T_1$. We deduce that for $t \ge T_1$, $\| x_i(t+1) - x_i(t) \| \ge 2\eps$ which proves that $x_i(t+1) \notin B(v, \eps)$. This is true for any particle located in $B(v, \eps)$ at time $t \ge T_1$. This proves that after time $T_1$, if $x_i(t+1) \in B(v, \eps)$, then $x_i(t) \notin B(v, \eps)$. We then conclude using Lemma~\ref{lem:cant_leave_vertices_neigh}.

            \item[(4)]The proof follows exactly as the one in~\cite[Lemma 4.4]{AA-GF-EZ:24} and is omitted.
    \end{itemize}
\end{proof}

\begin{figure}[h]
    \centering
    \subfigure[Situation of Remark \ref{rem:set_S}\label{fig:convergence_not_in_S_a}]{
        \includegraphics[width=0.6\textwidth]{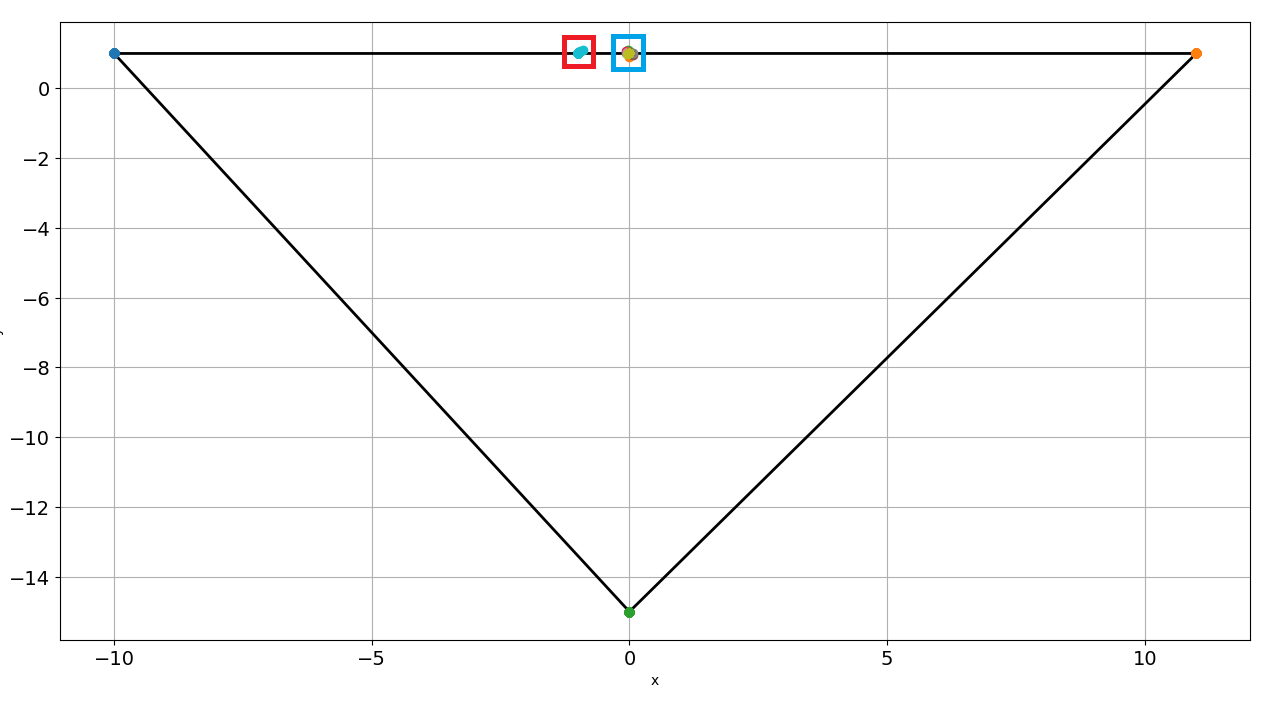}
    }

    \par\medskip
    \subfigure[Zoom on the red square from Figure \ref{fig:convergence_not_in_S_a}\label{fig:convergence_not_in_S_b}]{
        \includegraphics[width=0.45\textwidth]{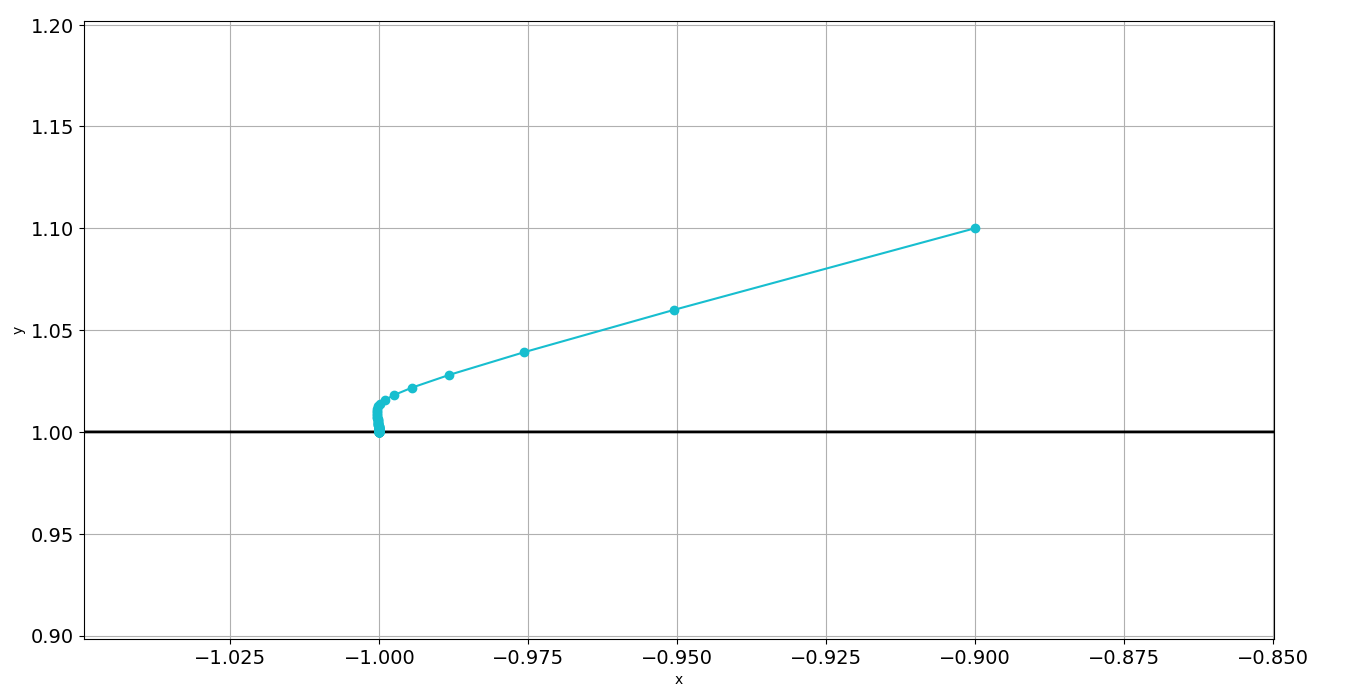}
    }
    \subfigure[Zoom on the blue square from Figure \ref{fig:convergence_not_in_S_a}\label{fig:convergence_not_in_S_c}]{
        \includegraphics[width=0.45\textwidth]{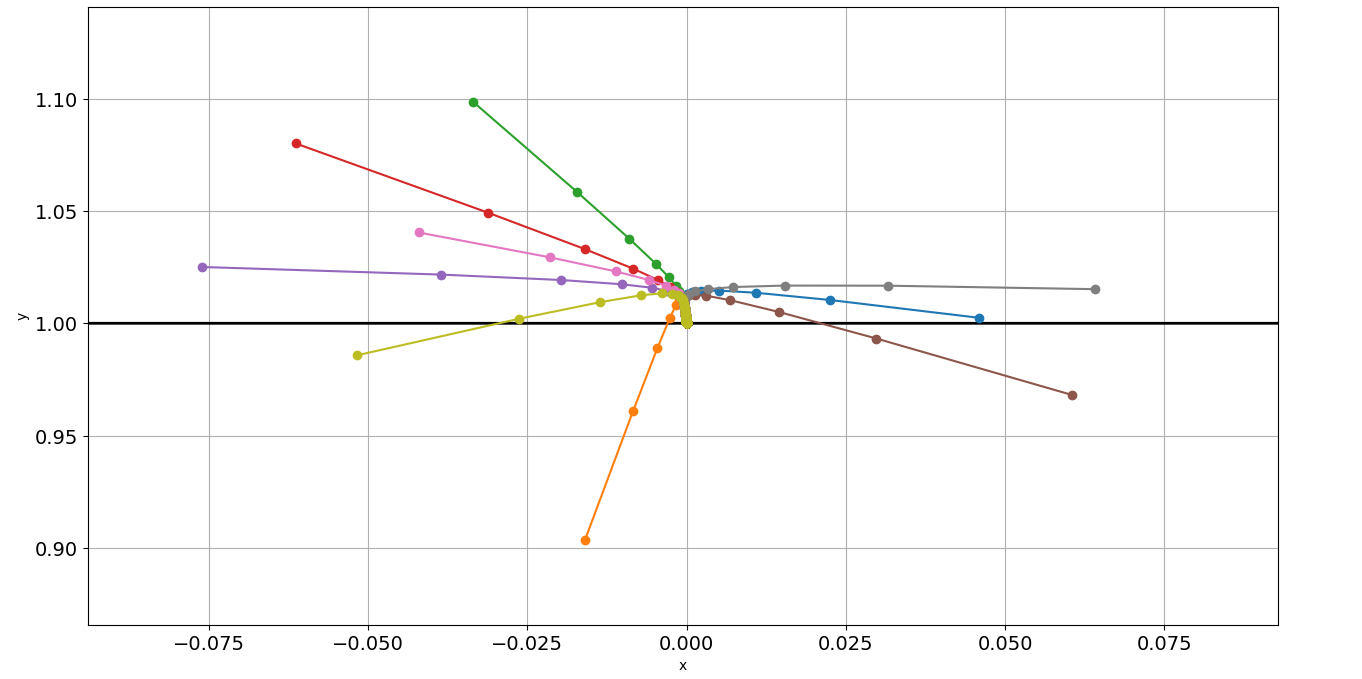}
    }
    \captionsetup{justification=centering}
    \caption{Situation of Remark \ref{ex:counter_example}. Figure (b) (resp (c)) is a zoom on the red (resp blue) square.}
    \label{fig:convergence_not_in_S}
\end{figure}

The upshot of Theorem~\ref{thm:S_carac} is that it provides a characterization for the maximal alignment set $ S $. We make an important remark on the relationship between the maximal alignment set and the $ \omega $-limit set of the dynamics~\eqref{eq:hardmax_delta}.
\begin{remark}\label{ex:counter_example}\longthmtitle{Convergence can happen outside the maximal alignment set}
    \label{rem:set_S}
    In~\cite{AA-GF-EZ:24} and~\cite{BG-CL-YP-PR:24}, the set $S$ is of interest, because it corresponds exactly to candidate set of convergence point, in that every particle converges to some point in $S$. This is not true for the dynamics~\eqref{eq:hardmax_delta}. We illustrate this with an example. Consider the following configuration
    \begin{itemize}
    \item One particle in $P_1 = (-10, 1)$, one in $P_2 = (11,1)$ and one in $P_3 =(0,-15)$; 
    \item Nine distinct particles randomly chosen in the square centered in $O = (0,1)$ with length $0.1$; 
    \item One particle in $D=(-0.9,1.1)$;
    \item $\delta = 8$.
    \end{itemize}
    Figure~\ref{fig:convergence_not_in_S} is an illustration of this configuration\footnote{Indeed, this numerical observation can easily be verified analytically.}.
    This situation is of interest because it shows that it is possible for tokens to converge to points that are not in $S$ (not a vertex nor the projection of the origin to some face of $K$, this is figure~\ref{fig:convergence_not_in_S_b}) in the non-trivial case (initial distribution not at equilibrium). In fact, this is the fact that $\delta$ is constant that allows convergence outside of $S$. Indeed, we can see that when we replace the constant $\delta$ by a sequence $(\delta(t))_{t \ge 0} \in [0, +\infty[^\mathbb{N}$ such that $\delta(t) \to 0$ as $t\to \infty$ (see section~\ref{sec:delta(t)}), we recover the convergence in $S$ exactly as in~\cite{AA-GF-EZ:24} and~\cite{BG-CL-YP-PR:24}.
\end{remark}

In spite of this remark, the inclusion $V \subset S$ is of importance, and will be used in our later constructions, c.f., Lemma~\ref{lem:empty_zone_near_vertices}.

\begin{lemma}
    \label{lem:norm_increasing}
    For any time $t \in \mathbb{N}$ and any $i \in [n]$, if $x_i(t) \in K$, $x_i(t) \ne 0$, and $\dist(x_i(t), \partial K) \ge \delta$, then $\|x_i(t+1)\|^2 - \| x_i(t) \|^2 \ge 2 \frac{\alpha}{1+\alpha} \frac{1}{n} \delta \|x_i(t) \| + \frac{\alpha^2}{(1+\alpha)^2} \frac{1}{n^2} \delta^2$.
\end{lemma}
\begin{proof}
    First of all, we claim that if $j \in C_i^\delta(t)$, then $x_j(t) \notin K$ or $\dist(x_j(t), \partial K) \le \delta$. Indeed, let 
    \[
    j_0 \in \argmax_{\ell \in [n]} \innerp{x_i(t)}{x_\ell(t)}\,.
    \]
    Using Theorem~\ref{thm:convex_hull_decreasing}, $x_{j_0}(t) \notin \mbox{int}(K)$ (because if $x_{j_0}(t) \in K$ then $x_{j_0}$ maximizes the linear function $x \mapsto \innerp{x_i(t)}{x}$ over the convex set $K$). This also shows that the strip $Z_i(t)$ defined in~\eqref{eq:Z_i(t)}, cannot be strictly included in $K$. Since this strip is of width $\delta$, we have that either $x_j(t) \notin K$, or 
    \[
    x_j(t) \in Z_i(t) \implies \dist(x_j(t), \partial K) \le \delta,
    \]
    proving our claim. 
     We now rewrite 
    \begin{equation*}
        \begin{aligned}
            \|x_i(t+1)\|^2 - \| x_i(t) \|^2 = 2 \frac{\alpha}{1+\alpha} &\frac{1}{|C_i^\delta(t)|} \sum_{j \in C_i^\delta(t)} \innerp{x_i(t)}{x_j(t) - x_i(t)}  \\
           &+ \Big\Vert \frac{\alpha}{1+\alpha} \frac{1}{|C_i^\delta(t)|} \sum_{j \in C_i^\delta(t)} x_j(t) - x_i(t) \Big\Vert^2\,. 
        \end{aligned}
    \end{equation*}
    Note that, using the observation above, $\innerp{x_i(t)}{x_j(t) - x_i(t)} \ge 0$. Indeed, we assumed that $\dist(x_i(t), \partial K) \ge \delta$, hence, for any $j \in C_i^\delta(t)$, $\innerp{x_i(t)}{x_j(t)} \ge \innerp{x_i(t)}{x_i(t)}$. Therefore, 
    \[
    \|\sum_{j \in C_i^\delta(t)} x_j(t) - x_i(t) \| \ge \sum_{j \in C_i^\delta(t)}\frac{\innerp{x_i(t)}{x_j(t) - x_i(t)}}{\|x_i(t)\|} \ge \frac{1}{\|x_i(t)\|} \innerp{x_i(t)}{x_{j_0}(t) - x_i(t)}
    \]
    and
    \begin{equation}
    \begin{aligned}
        \|x_i(t+1)\|^2 - \| x_i(t) \|^2 &\ge 2 \frac{\alpha}{1+\alpha} \frac{1}{n} \innerp{x_i(t)}{x_{j_0}(t) - x_i(t)}\\ &\qquad+ \frac{\alpha^2}{(1+\alpha)^2} \frac{1}{n^2} \frac{1}{\|x_i(t)\|^2} \innerp{x_i(t)}{x_{j_0}(t) - x_i(t)}^2 \\
        &\ge 2 \frac{\alpha}{1+\alpha} \frac{1}{n} \delta \|x_i(t) \| +\frac{\alpha^2}{(1+\alpha)^2} \frac{1}{n^2} \delta^2.
    \end{aligned}
    \end{equation}
    This yields the result. 
\end{proof}
\begin{proposition}
    \label{prop:center_area_empty}
    Let  
    \[
    \nu = \frac{\alpha^2}{(1+\alpha)^2} \frac{1}{n^2} \delta^2, \quad M := \max_{\ell \in [n]} \| x_\ell \|_{L^\infty}, \quad \mathrm{and}  \quad m := \min_{\ell \in [n]} \| x_\ell(0) \|\,.
    \]    
   
    Suppose that the following assumptions holds true:
    \begin{enumerate}
        \item $0 \in K$;
        \item $\dist(0, \partial K) \ge \delta$;
        \item The initial tokens are all non zero;
    \end{enumerate}
    Then, for any $i \in [n]$ and $t \ge \subscr{t}{lim} := \lfloor\frac{2c^\star(M+\nu)}{m\nu} \rfloor + 1$, we have that $ \|x_i(t)\| > c^\star$ where    
    \[
    c^\star:= \sup \{c > 0 \; | \;\dist(x, \partial K) \ge \delta \quad  \forall x \in K\cap B(0, c) \}.
    \]
\end{proposition}
\begin{proof}
    First, note that by our assumption~(3), $ m>0$. 
    Using Lemma~\ref{lem:norm_increasing}, we know that any particle $x_i(t) \in B(0, c^\star)\backslash \{0\}$ satisfies $\|x_i(t+1)\|^2 - \| x_i(t) \|^2 \ge  \frac{\alpha^2}{(1+\alpha)^2} \frac{1}{n^2} \delta^2$. Therefore :
    \begin{equation*}
        \begin{aligned}
            \| x_i(t+1)\| &\ge \| x_i(t) \| \sqrt{ 1 + \frac{\nu}{\|x_i(t)\|}} \\
            &\ge \| x_i(t)\| \sqrt{1 + \frac{\nu}{M}} \\
            &\ge \| x_i(t) \|\left( 1 + \frac{1}{2}\frac{\nu}{M + \nu}\right) \\
            &\ge \| x_i(t) \| + \frac{1}{2} \frac{m\nu }{M + \nu}\,.
        \end{aligned}
    \end{equation*}
    Hence, for $t \ge \subscr{t}{lim} = \lfloor\frac{2c^\star(M+\nu)}{m\nu} \rfloor + 1$, we have that $\|x_i(t)\| > c^\star$ for any $i \in [n]$, as claimed. 
\end{proof}

\section{Micro-level properties of the trajectories}\label{sec:micro}

In the last section, we made an important observation that the maximal alignment set of~\eqref{eq:hardmax_delta} does not fully characterize where the tokens eventually converge to. This motivates further micro-level studies to gain insights about the behavior of the trajectories of the tokens in regions near the vertices of $ K $. In particular, we are interested in capturing an invariance-type property which we precisely define next.
\begin{definition}\longthmtitle{Quiescent sets}
\label{def:quiescent_set}
    We say that a set $ W \subset \real^d$ is a \emph{quiescent set} for the dynamics~\eqref{eq:hardmax_delta} if there exists a finite time $ T >0 $ such that, if token $ x_i(T) \in W $, then $ x_i(t) \in W $ for all $ t \geq T $, and vice versa,
        if token $ x_i(T) \notin W $, then $ x_i(t) \notin W $ for all $ t \geq T $. In this case, we refer to $ T $ as the \emph{settling time} for the quiescent set $ W$. 
\end{definition}
In this sense, the set of tokens in a quiescent set for the dynamics eventually will become invariant with time. Let us note that since there is only a finite number of particles, the reciprocal statement in Definition~\ref{def:quiescent_set} is redundant. We let it as is for clarity.  We now state our main result in this section. 
\begin{theorem}\longthmtitle{Quiescent sets near vertices of $ K $}
\label{thm:convergence_to_vertices}
For $v \in V\backslash \{0\}$ and $0<\varepsilon\leq\frac{\delta}{2+2\alpha}$, 
where $ \delta $ is the alignment sensitivity parameter, the set $B(v, \varepsilon)$ is a quiescent set for the dynamics~\eqref{eq:hardmax_delta}. Moreover, there exists a settling time $T>0$ such that any token in $B(v, \varepsilon)$ after $T$ converges to $v$ with a geometric rate.
\end{theorem}

The proof of this result is involved and relies on several intermediate steps, which we develop next.
From now on, for any $v \in V$, we will refer to the set 
\begin{equation}
\label{eq:zone_of_influence}
    \mathcal{A}_v(t) := \{x \in K_{\eta(t)} \; | \; \innerp{v-x_i(t)}{v} \le \delta \|v \|\}
\end{equation}
as the \textit{zone of influence of $v$}. This name is motivated by the fact that if a particle is located in the geometric position $v$ at time $t$, then it will be influenced by all the particles located in $\mathcal{A}_v(t)$.

Our first result, roughly speaking, states that for $t$ large enough, a particle cannot move from outside of $\mathcal{A}_v(t)$ to inside $B(v, \varepsilon)$ in just one iteration.

\begin{lemma}
\label{lem:4.1}
    Let $0 < \varepsilon < \frac{\delta}{1+\alpha}$.  There exists a time $T \in \mathbb{N}$ such that, for any $v \in V\backslash \{0\}$, any $t \ge T$, and any $i \in [n]$, if $\innerp{v-x_i(t)}{v} \ge \delta \| v\|$ then $\| x_i(t+1) - v \| > \eps$.
    
\end{lemma}
\begin{proof}
    We proceed by contradiction. Suppose we can find a vertex $v$ particle $x_i$ satisfying :
    \begin{itemize}
        \item $\innerp{v - x_i(t)}{v} \ge \delta \| v \|$
        \item $\| x_i(t+1) - v \| \le \eps$
    \end{itemize}
    for $t$ arbitrarily large. \\
    Then we can write :
    \begin{equation}
    \label{eq:bound1}
        \begin{aligned}
            \innerp{x_i(t+1) - x_i(t)}{v} &= \innerp{x_i(t+1)-v}{v} + \innerp{v - x_i(t)}{v} \\
            &\ge (\delta - \eps) \|v \|\,.
        \end{aligned}
    \end{equation}
    But we also have :
    \begin{equation}
    \label{eq:bound2}
        \begin{aligned}
            \innerp{x_i(t+1) - x_i(t)}{v} &= \frac{\alpha}{1+\alpha} \frac{1}{|C_i^\delta(t)|} \sum_{j \in C_i^\delta(t)} \innerp{x_j(t) - x_i(t)}{v} \\
            &\le \frac{\alpha}{1+\alpha} \max_{j \in C_i^\delta(t)} \innerp{x_j(t) - x_i(t)}{v} \\
            &\le \frac{\alpha}{1+\alpha} ( \delta + \eta(t) ) \| v\|\,.
        \end{aligned}
    \end{equation}
    The last inequality holds thanks to Theorem~\ref{thm:S_carac}, item 3. Indeed, since for any $j \in [n]$, $x_j(t) \in K_{\eta(t)}$, there exists $p(t) \in K$ such that $\| x_j(t) - p(t) \| \le \eta(t)$. Therefore:
    \begin{equation*}
        \begin{aligned}
            \innerp{x_j(t) - v}{v} &= \langle x_j(t) - p(t), v\rangle + \langle p(t)-v,v\rangle \\
            &\le \eta(t) \|v\| + \langle p(t), v \rangle - \max_{x \in K} \innerp{x}{v} \\
            &\le \eta(t) \|v\|\,.
        \end{aligned}
    \end{equation*}
    Combining equations~\eqref{eq:bound1} and~\eqref{eq:bound2}, we get 
    \begin{equation}
        \label{eq:lower_bound_eps}
        \frac{\delta - \alpha\eta(t) }{1+\alpha} \le \eps\,.
    \end{equation}
    Since $\eta(t) \longrightarrow 0$ as $t \rightarrow +\infty$, if there exists an unbounded sequence of time such that~\eqref{eq:lower_bound_eps} holds, we would have $\varepsilon \ge \frac{\delta}{1+\alpha}$ which contradicts our assumption. Therefore, there exists $T(\varepsilon)\in \mathbb{N}$ after which~\eqref{eq:lower_bound_eps} does not hold which conclude the proof.
\end{proof}
Our next result states that, for $v \in V$ and $t$ large enough, the only particles in the zone of influence of $v$ must be close to $v$.
\begin{lemma}
\label{lem:empty_zone_near_vertices}
     Let $v \in V \backslash \{0\}$, $\varepsilon > 0$ and $0 < c < \delta$. There exists a time $T \in \mathbb{N}$ such that, for any $i \in [n]$ and $t \ge T$, if $\langle v, v-x_i(t) \rangle \leq (\delta - c ) \lVert v \rVert$, then $\lVert x_i(t) - v \rVert \leq \varepsilon$.
\end{lemma}
\begin{proof}
    Suppose that we can find arbitrarily large time $t$ such that $\langle v, v-x_i(t) \rangle \leq (\delta - c )\lVert v \rVert$. We want to prove that this implies that $x_i(t)$ influences any particle close enough to $v$. 
    \begin{align*}
    \dmin = \min_{w \in V\backslash \{0\}} \|w\| > 0,\quad  \dmax = \max_{w \in V} \| v_k \|, \quad \text{and} \quad M = \max_{\ell \in [n]} \| x_\ell\|_{L^{\infty}} < + \infty\,.
    \end{align*}
    We also define
    \begin{equation*}
        0 <\mu \le \min \left( c, \frac{\dmin}{2}\right)\,.
    \end{equation*}
    Let $T$ such that $\eta(t) \le \mu$ for any $t \ge T$. We claim that for any $t \ge T$ such that $\langle v, v-x_i(t) \rangle \leq (\delta - c )\lVert v \rVert$, $x_i(t)$ influences every particle in $B(v, \mu)$. Indeed, if $x_j(t) \in B(v, \mu)$, then the assumption on $\mu$ gives that $\|x_j(t) \| \ge \frac{\dmin}{2}$ and using Lemma~\ref{lem:diff_max} and Cauchy-Schwartz inequality, we have that
    \begin{equation*}
        \begin{aligned}
            \max_{\ell \in [n]} \innerp{x_j(t)}{x_\ell(t)}& - \innerp{x_j(t)}{x_i(t)}\\ & \le \eta(t) \|x_j(t)\| + \max_{k \in [r]} \innerp{x_j(t)}{v_k} - \innerp{x_j(t)}{x_i(t)} \\
            &\le \mu \|x_j(t) \| + \mu \cdot \dmax + \mu \cdot \| x_i(t) \| + \max_{k \in [r]} \innerp{v}{v_k} - \innerp{v}{x_i(t)} \\
            &\le \mu \| x_j(t) \| + \mu \left( \dmax + M\right) + \innerp{v}{v - x_i(t)}.
        \end{aligned}
    \end{equation*}
    The last inequality holds because we know that $v \in S$ (Theorem~\ref{thm:S_carac}). Now, we use the fact that $\innerp{v}{v - x_i(t)} \le (\delta - c ) \|v \|$ by assumption. It follows :
    \begin{equation*}
        \begin{aligned}
            \max_{\ell \in [n]} \innerp{x_j(t)}{x_\ell(t)}& - \innerp{x_j(t)}{x_i(t)} \\ &\le \mu \| x_j(t) \| + \mu \left( \dmax + M\right) + (\delta - c )( \| x_j(t) \| + \mu ) \\
            &\le \delta \| x_j(t) \| - (c-\mu) \| x_j(t) \| + \mu \left( \dmax + M\right) + (\delta - c ) \mu  \\
            & \le \delta \| x_j(t) \| - (c - \mu)\frac{\dmin}{2} + \mu \left( \dmax + M\right) + (\delta - c ) \mu\,.
        \end{aligned}
    \end{equation*}
    Note that the quantity $- (c - \mu)\frac{\dmin}{2} + \mu \left( \dmax + M\right) + (\delta - c ) \mu$ tends to $-\frac{c\dmin}{2}$ when $\mu \rightarrow 0$. Therefore, there exists $\mu_0 >0$ such that if $0<\mu\le \mu_0$ and $T$ is such that $\eta(t) \le \mu$ for $t\ge T$, then $\max_{\ell \in [n]} \innerp{x_j(t)}{x_\ell(t)} - \innerp{x_j(t)}{x_i(t)} \le \delta \|x_j(t)\| $ for any $t\ge T$.
    Hence, $i \in C_j^\delta(t)$ for any $t \ge T$. Note that $T$ depends only on the fixed parameters $\dmin$, $\dmax$, $M$, $\delta$ and $c$.
    
    Recall now the dynamics:
    \begin{equation}
    \label{eq:proof_lem_empty_near_vertices}
        x_j(t+1) - x_j(t) = \frac{\alpha}{1 + \alpha} \frac{1}{| C_j^\delta(t) |}\left( x_i(t) - x_j(t) + \sum_{k \in C_j^\delta(t)\backslash \{i\} }(x_k(t) - x_j(t)) \right)\,.
    \end{equation}
    Suppose by way of contradiction that $\|x_i(t) - v\| \ge \eps$.
    We define the unit vector $\mathbf{n} = \frac{x_i(t)-v}{\|x_i(t) - v\|}$. We have : $\innerp{x_i(t)-x_j(t)}{\mathbf{n}} \ge \eps - \mu$ and $\innerp{x_k(t)-x_j(t)}{\mathbf{n}} \ge -2\mu$ for any $k \in C_j^\delta(t)$. Therefore :
    \begin{equation*}
        \begin{aligned}
            \| x_j(t+1) - x_j&(t) \|^2 \ge \innerp{x_j(t+1)-x_j(t)}{\mathbf{n}} \\
            &= \frac{\alpha}{1 + \alpha} \frac{1}{| C_j^\delta(t) |}\left( \innerp{x_i(t) - x_j(t)}{\mathbf{n}} + \sum_{k \in C_j^\delta(t)\backslash \{i\} }\innerp{x_k(t) - x_j(t)}{\mathbf{n}} \right) \\
            &\ge \frac{\alpha}{1 + \alpha} \frac{1}{n} \left( \eps - \mu - 2(n-1)\mu \right)\,. 
        \end{aligned}
    \end{equation*}
    Again, as $\mu\rightarrow 0$, the last term tends to $\frac{\alpha}{1+\alpha}\frac{\eps}{n}$. Therefore, there exists $0 <\mu_1 < \mu_0$ such that if $T$ is such that $\eta(t) \le \mu_1$ for all $t \ge T$, then $\|x_i(t+1) - x_i(t) \|^2 \ge 2\mu_1$. This implies that $x_j(t)$ leaves the $\mu-$neighborhood of $v$ at time $t+1$. Note that this is true for any particle located in $B(v, \mu)$ at time $t\ge T$. However, Lemma~\ref{lem:cant_leave_vertices_neigh} implies that this is impossible. 
\end{proof}
We are now in a position to prove Theorem~\ref{thm:convergence_to_vertices}. 
\begin{proof}[Proof of Theorem~\ref{thm:convergence_to_vertices}]
    We divide the proof into three parts; a rough description is added to help guide the reader: 
    
    \textbf{Particles outside cannot enter $B(v, \varepsilon)$:} This follows from Lemma~\ref{lem:4.1}.
    Note that the assumption made on $\varepsilon$ is $\varepsilon \leq \frac{\delta}{2 + 2\alpha}$. The reason is that we can make the same reasoning as in Lemma~\ref{lem:4.1} but with $d \ge \frac{\delta}{2}$ instead of $d \ge \delta$. Therefore, with this choice of $\varepsilon$, if $\lVert x_i(t) - v \Vert \ge \frac{\delta}{2}$, then $\lVert x_i(t+1) - v \rVert > \varepsilon$ for $t$ larger than some $T$. The consequence is that if there exists $T \leq t_1 < t_2$ such that $x_i(t_1) \notin B(v, \varepsilon)$ but $x_i(t_2) \in B(v, \varepsilon)$, there must exists $t_1 \leq t_3 < t_2$ such that $x_i(t_3) \in B(v, \frac{\delta}{2}) \backslash B(v, \varepsilon)$. This is impossible after a certain time using Lemma~\ref{lem:empty_zone_near_vertices} with $c = \frac{\delta}{2}$.\\
    
    \textbf{Particles in $B(v, \varepsilon)$ cannot leave:}. This is a logical consequence of the previous item: There is only a finite number of particles and we know that after $T$ defined in item 1, particles outside $B(v, \eps)$ cannot enter this set. Therefore, the process of a particle leaving $B(v, \eps)$ can only happens finitely many times. Upon increasing $T$ we can suppose that no particle leave $B(v, \eps)$ after $T$.\\
    
    \begin{sublemma}\longthmtitle{Convergence rate of token in the quiescent set is geometric:}
        \label{sublem:neigh_localized_arnoud_vertices}
        Let $0<\eps\le \frac{\delta}{2 + 2\alpha}$ and let $v \in V \backslash \{0\}$. There exists a time $t^{*} \in \mathbb{N}$ such that, for any $t \ge t^{*}$ and for any $i \in [n]$ such that $x_i(t) \in B(v, \eps)$, if $j \in C_i^\delta(t)$, then $x_j(t) \in B(v, \eps)$.
    \end{sublemma}
    \begin{proof}
    If $x_i(t) \in B(v, \eps)$, it follows $x_i(t) \in \mathcal{A}_v(t)$ where $\mathcal{A}_v(t)$ is the zone of influence of $ v(t)$, defined in~\eqref{eq:zone_of_influence}. Hence, applying Lemma~\ref{lem:empty_zone_near_vertices} with $c = \delta - \eps >0$, we know that for any $\xi >0$, there exists $T(\xi) \ge 0$ such that $B(v, \eps) \backslash B(v, \xi)$ contains no particle for any time $t \ge T(\xi)$. Then, using similar lines of reasoning as in Theorem~\ref{thm:S_carac} item 3, we know that for $\xi$ small enough, if $x_i(t) \in B(v, \xi)$, $\|x_j(t) - v \| \ge \eps$ and $j \in C_i^\delta(t)$, then $x_i(t+1) \notin B(v, \xi)$. Take such a $\xi > 0$ and consider the associated time $T(\xi)$. We also consider the time $T_1$ as defined in the two previous items after which the particles in $B(v, \xi)$ are fixed (they can't leave and no other particle can come inside). We define $t^{*} = \max(T_1, T(\xi))$ and we claim that $t^{*}$ satisfies the statement of this sublemma. Indeed, let $t \ge t^{*}$ and $i \in [n]$ such that $x_i(t) \in B(v, \eps)$. Since $t \ge T(\xi)$, $B(v, \eps) \backslash B(v, \xi)$ contains no particle so necessarily, $x_i(t) \in B(v, \xi)$. Now, if there exists an index $j \in C_i^\delta(t)$ satisfying $x_j(t) \notin B(v, \eps)$, then we would have $x_i(t) \notin B(v, \xi)$ which is impossible because $t \ge T_1$. Therefore, $x_j(t) \in B(v,\eps)$ which conclude the proof.
    \end{proof}    
    Using Sublemma~\ref{sublem:neigh_localized_arnoud_vertices}, we can define $0<\eps '\le\eps$ and suppose that any particles in $B(v, \eps ')$ is influenced exactly by all the other particles in $B(v, \eps')$. Upon applying items 1 and 2 for $\eps'$, we suppose that $T$ is large enough so that particles in $B(v, \eps')$ are fixed. Let us denote by $x_1, ..., x_p$, where $ p $ is some positive integer, the particles in $B(v, \eps')$. Let 
    $
    m(t) = \frac{1}{p}\sum_{j=1}^px_j(t)
    $
    be the center of mass of the particles. We have that
    \[
    x_j(t+1) = \frac{1}{1+\alpha}x_j(t) + \frac{\alpha}{1+\alpha}m(t).
    \]
    Therefore, $m(t+1) = m(t)$: the center of mass is constant. As a result, we necessarily have $m(t) = v$, because we know that we can find particle arbitrarily close to $v$ for $t$ large enough and $m(t)$ is constant. Now, let 
    \begin{equation}\label{eq:D-mass}
    D(t) = \max_{1\leq j \leq p} \lVert x_j(t) - m(t) \rVert.
        \end{equation}
    It is easy to show that $D(t+1) = \frac{1}{1+\alpha}D(t)$, namely we have a geometric decrease of the maximal distance between the particles in $B(v, \eps')$ and $m(t) = v$. 
\end{proof}
The next result is a straightforward consequence of the the geometric convergence result in Theorem~\ref{thm:convergence_to_vertices},
and in fact is similar to what is observed in~\cite[Lemma 3.1]{AA-GF-EZ:24} for the dynamics~\eqref{eq:pure_hardmax}. This phenomenon distinguishes the transformer attention dynamics from classical opinion dynamics such as Hegselmann-Krause, which have finite-time convergence properties~\cite{AN-BT:2011,SRE-TB-AN-BT:2013}.
\begin{corollary}\longthmtitle{Lack of finite-time convergence}
    \label{cor:no_finite_time_convergence}
    If the initial distribution $(x_i(0))_{1\le i \le n}$ is not at equilibrium, then there is no finite time convergence for dynamics~\eqref{eq:hardmax_delta}.
\end{corollary}

We refer the reader to Section~\ref{sec:future-sim}, where we demonstrate the results of this section through some case studies.

\section{Localmax dynamics in the vanishing alignment sensitivity regime}
\label{sec:delta(t)}

In that section, we consider a natural modification of the localmax dynamics~\eqref{eq:neigh-localmax},
\begin{equation}
    C_i^\delta(t) = \{ j \in [n] \; | \; \max_{k \in [n]} \langle Ax_i(t), x_k(t) \rangle - \langle A x_i(t), x_j(t) \rangle \leq \delta(t) \lVert Ax_i(t) \rVert \} 
    \label{eq:neigh_localmax_delta(t)}
\end{equation}
where $(\delta(t))_{t \ge 0}$ is a sequence of non-negative real numbers satisfying $\delta(t) \to 0$ as $t\to \infty$. 
Note that as $ \delta(t)$ approaches zero, this dynamics recovers the one~\eqref{eq:neigh-hardmax}. We prove the following result.
\begin{theorem}
\label{thm:convergence_localmax_delta(t)}
    Let $((x_i(t))_{i\in[n]})_{t \ge 0}$ be a sequence of tokens following dynamics \eqref{eq:hardmax_delta} with the neighborhood defined in \eqref{eq:neigh_localmax_delta(t)} for some  arbitrary initial distribution $(x_i(0))_{1\le i\le n}$. Then, for any $i \in [n]$, there exists $w \in S$ such that $x_i(t) \xrightarrow[t \rightarrow+\infty]{}w$.
\end{theorem}
The upshot of this result is that, in contrast to Theorem~\ref{thm:convergence_to_vertices}, as $ \delta(t) $ becomes small, the maximal alignment set~\eqref{eq:maximal-alig-S} fully characterizes where the trajectories converge to.
As the first step in the proof, we show the following result. Naturally, after adjusting the terminology, the result recovers~\cite[Theorem~1.1]{AA-GF-EZ:24} as a corollary.
\begin{lemma}
    \label{lem:lower_bound_outside_of_Seps}
    For any $\eps >0$, we define the set $S_\eps$ containing all the points located at a distance of at most $\eps$ from $S$:
    \begin{equation}
        \label{eq:set_S_eps}
        S_\eps = \{ x \in \mathbb{R}^d \; | \; \exists w \in S, \; \| x - w \| \le \eps \}.
    \end{equation}
    Let $\eps > 0 $, and consider the dynamics~\eqref{eq:hardmax_delta},     
    with the neighborhood $C_i^\delta(t)$ as defined in~\eqref{eq:neigh_localmax_delta(t)}, with $ \delta(t)$ a vanishing sequence of alignment sensitivity parameter. 
    There exists a constant $\gamma > 0$ and a time $T > 0$ such that, for any $t \ge T$, if $x_i(t) \notin S_\eps$ then $\|x_i(t+1)\|^2 - \| x_i(t) \|^2 \ge \gamma \eps.$
\end{lemma}
Clearly, the statement is related to the Lemma~\ref{lem:norm_increasing}, but crucially, the result is concerned with a vanishing time-varying alignment sensitivity parameter $\delta(t)$. Moreover, as outlined below, the proof additionally uses some results from~\cite{BG-CL-YP-PR:24}.
\begin{proof}
    Recall from the proof of Lemma~\ref{lem:norm_increasing} that
\begin{equation}
    \|x_i(t+1)\|^2 - \| x_i(t) \|^2 \ge 2 \frac{\alpha}{1+\alpha} \frac{1}{|C_i^\delta(t)|} \sum_{j \in C_i^\delta(t)} \innerp{x_i(t)}{x_j(t) - x_i(t)}
\end{equation}
and set $a_{ij}(t) = \innerp{x_i(t)}{x_j(t) - x_i(t)}$. We need to find a bound on $a_{ij}(t)$ for all $j \in C_i^\delta(t)$. We start by fixing an $\eps_0 > 0$ such that all the $\eps_0$-neighborhood of points in $S$ are separated by a distance of at most $\eps_0$. Then, we use two results from~\cite{BG-CL-YP-PR:24}:
\begin{enumerate}
    \item There exists $\lambda > 0$ such that, if $x \in K_\lambda \backslash S_{\eps_0}$ then, using 
    the proof of the first case of~\cite[Theorem 3.1, Claim 2]{BG-CL-YP-PR:24},
    $\max_{k \in [r]} \langle x, v_k \rangle - \innerp{x}{x} \ge c$ with $c >0$. We now define $T_1\in \mathbb{N}$ such that $\eta(t) \le \lambda$ for any $t \ge T_1$. After time $T_1$, all the particles are in $K_\lambda$.
    \item There exists $\beta >0$ such that for any $\omega \in S$, if $x \in K \cap B(\omega, \eps_0)$, then, using \cite[Lemma 8.4] {BG-CL-YP-PR:24},
    \begin{equation}
        \max_{k \in [r]} \innerp{x}{v_k - x} \ge \beta \|x - \omega \|\,.
    \end{equation}
    Following the same steps as in~\cite[Theorem 3.1, Claim 2]{BG-CL-YP-PR:24}, we conclude that there exists a time $T_2 \ge 0$ such that for $t \ge T_2$, if $x_i(t) \in S_{\eps_0} \backslash S_\eps$, then $\max_{k \in [r]} \innerp{x_i(t)}{v_k - x_i(t)} \ge \frac{\beta}{2}\eps$.
\end{enumerate}
Therefore, for any $0 < \eps \le \frac{2c}{\beta}$, any $t \ge \max(T_1, T_2)$, and $x_i(t) \notin S_\eps$, by combining the two statements above, we have that 
\begin{equation}
    \max_{j \in [n]}a_{ij}(t) \ge \max_{k \in [r]} \innerp{x_i(t)}{v_k - x_i(t)} \ge \Tilde{\gamma}\eps
\end{equation}
with $\Tilde{\gamma} = \frac{\beta}{2}$. Now define $T_3 \ge \max(T_1, T_2)$ such that, for any $t\ge T_3$, $\delta(t) < \frac{\Tilde{\gamma} \eps}{M}$ where $M = \max_{\ell \in [n]}\| x_\ell \|_{L^\infty}$. Let $t \ge T_3$. We have that 
\begin{equation*}
    \begin{aligned}
        j \in C_i^\delta(t) &\iff \max_{\ell \in [n]} a_{il}(t) - a_{ij}(t) \le \delta(t) \| x_i(t) \| 
        \implies \Tilde{\gamma} \eps - a_{ij}(t) \le \Tilde{\gamma}\eps,
    \end{aligned}
\end{equation*}
which implies that $a_{ij}(t) \ge 0$. Therefore, 
\begin{equation}
    \|x_i(t+1)\|^2 - \| x_i(t) \|^2 \ge 2 \frac{\alpha}{1+\alpha} \frac{\Tilde{\gamma} \eps}{n}\,.
\end{equation}
\end{proof}
We are now able to proceed with the proof the Theorem \ref{thm:convergence_localmax_delta(t)}.
\begin{proof}[Proof of Theorem~\ref{thm:convergence_localmax_delta(t)}]
    Our goal is to prove that if $\eps >0$, all the particles are located in $S_\eps$ after a finite time (in fact, we only need to prove this for $\eps$ small enough). We first discuss the case where $0 \in S$, and then treat the case where $\| x_i(t)\|$ is non zero.
    
    \textbf{The case $0\in S$:} Let $\eps >0$. According to Lemma~\ref{lem:lower_bound_outside_of_Seps}, there exists $\Tilde{T} \in \mathbb{N}$ such that for any $t \ge \Tilde{T}$ and any $i \in [n]$, if $x_i(t) \notin B(0, \eps)$, then it will never enter it. We can therefore distinguish between particles that remain indefinitely in $B(0, \eps) \subset S_\eps$ for which nothing further needs to be proved, and particles that eventually leave $B(0, \eps)$ whose norm is then bounded below by $\eps$. With this at hand, we assume from now on that $\| x_i(t)\|$ is non zero. We first gather some preliminary observations:
    
    \textbf{Preliminary results:}
    Let $w \in S$. By definition, $\innerp{w}{v_k} = \max_{\ell \in [r]} \innerp{w}{v_\ell}$ for any $k \in \mathcal{I}_w$. Let us denote by $F_w$ the face of $K$ generated by the vertices in $\mathcal{I}_w$. 
    We define 
    \[
    F_w^\eps = \{ x \in \mathbb{R}^d \; | \; \dist(x, F_w) \le \eps \}.
    \]
    For $x \in K_{\eta(t)}$, define $\mathcal{I}_x := \{ k \in [r] \; | \; \innerp{x}{v_k} = \max_{\ell \in [r]} \innerp{x}{v_\ell}\}$. Let $\eps_0$ and $c>0$ be as defined in Lemma~\ref{lem:max_inner_prod_near_vertices}. This lemma gives that for any $0 < \eps \le \eps_0$ and any $x \in F_w^\eps$, $\mathcal{I}_x \subset \mathcal{I}_w$ which implies that the face $F_x$ of $K$ generated by the vertices in $\mathcal{I}_x$ is included in $F_w$. 
    
    Next, for $x \in K_{\eta(t)}$, we define the hyperplanes 
    \begin{align*}
    H_1 :=& \{ y \in \mathbb{R}^d \; | \; \innerp{x}{y} = \max_{\ell \in [r]} \innerp{x}{v_\ell}\}, \quad  \mathrm{and} \\
    H_2 :=& \{ y \in \mathbb{R}^d \; | \; \innerp{x}{y} = \max_{\ell \in [r]} \innerp{x}{v_\ell} - \delta(t) \|x\|\}.
    \end{align*}
    
    For any $x \in K_{\eta(t)} \backslash B(0, \eps)$, let $\mathcal{L}(x,t)$ be the strip of width $\delta(t)$ 
    delimited by the hyperplanes $H_1$ and $H_2$. By definition, $H_1$ is tangential to $K$ in $F_x$ and only in $F_x$. Thus, $H_1 \cap K = F_x \subset F_w$ for any $x \in F_w^\eps$. 
    This implies that, for any $\lambda >0$
    there exists $a(\lambda) > 0$ (a scalar depending on $\lambda$ and on the geometry of $K$ but not on $x$) such that if $\delta(t) \le a(\lambda)$, then for any $x \in F_w^\eps$, 
    \[
    y \in \mathcal{L}(x, t)\cap K \implies \dist(y, F_w)\le \lambda.
    \]
    Therefore, there exists $T(\lambda) \ge 0$ such that if $t\ge T(\lambda)$ then $\delta(t) \le a(\lambda)$ and thus,
    \begin{equation}
    \label{eq:implication_Zi}
        y \in \mathcal{L}(x, t) \cap K \implies \dist(y, F_w) \le \lambda, \;\;\; \forall x \in F_w^\eps\,.
    \end{equation}
    Furthermore, we know that if $x_i(t) \in F_w^\eps$ then
    \begin{equation}
    \label{eq:inclusion_Zi}
        Z_i(t) \cap K \subset \mathcal{L}(x_i(t), t) \cap K.
    \end{equation}
    This is because if $y \in K$ is such that $\max_{\ell \in [n]} \innerp{x_i(t)}{x_\ell(t)} - \innerp{x_i(t)}{y} \le \delta(t) \| x_i(t)\|$, then using the left inequality of Lemma \ref{lem:diff_max}, it follows $\innerp{x_i(t)}{y} \ge \max_{\ell \in [r]} \innerp{x_i(t)}{v_\ell} - \delta(t)\|x_i(t)\|$ and so $y \in \mathcal{L}(x_i(t), t) \cap K$. 
    In what precedes, $\eps$ and $a(\lambda)$ depend on $w\in S$ (actually, on the local geometry of $K$ around $F_w$). However, since $S$ is finite, we can assume that the same two quantities are valid for any $w \in K$.  We fix $\eps >0$ small enough so that the previous statements hold for any $w \in S$.
    \begin{sublemma}\label{sublemma:1}
         For any $\lambda > 0$, if $t \ge T(\lambda)$ and $x_i(t) \in F_w^\eps$, then $$| 2\lambda - \dist (x_i(t), F_w) |\to 0 \quad\text{ as } \quad t\to\infty\,.$$
    \end{sublemma}
    \textbf{Proof of Sublemma~\ref{sublemma:1}:} Let $w \in S$ and $\lambda > 0$. If $x_i(t) \in F_w^\eps$ then for any $j \in C_i^\delta(t)$, we have $x_j(t) \in Z_i(t) \cap K_{\eta(t)}$. If $x_j(t) \in K$ then~\eqref{eq:inclusion_Zi} gives that $x_j(t) \in \mathcal{L}(x_i(t), t)$ and~\eqref{eq:implication_Zi} ensures that $\mbox{dist}(x_j(t), F_w) \le \lambda$. If $x_j(t) \in Z_i(t)\cup K_{\eta(t)} \backslash K$, then~\eqref{eq:inclusion_Zi} gives $\dist(x_j(t), \mathcal{L}(x_i(t), t)) \le \eta(t)$ and~\eqref{eq:implication_Zi} gives $\dist(x_j(t), F_w) \le \lambda + \eta(t)$. Upon increasing $T(\lambda)$, we can assume that if $t\ge T(\lambda)$, $\eta(t) \le \lambda$. To summarize, if $t \ge T(\lambda)$ and $x_i(t) \in F_w^\eps$, then for any $j \in C_i^\delta(t)$, $\dist(x_j(t), F_w) \le 2\lambda$, which yields the claims.
    \oprocend  
    
    \textbf{Main part of the proof :} First, note that for any $w \in S$, since $w$ is the orthogonal projection of the origin to $F_w$, we have $\sup_{y \in B(w, \eps)} \|y \|^2 = \|w\|^2 + \eps^2$. Also, for the same reason,     
    \[
    \inf \{ \|x\|^2, x \notin B(w, \eps) \; \mbox{and} \; \dist(x, F_w) \le c \|w\|\} > (1-c)^2\|w \|^2 + \eps^2.
    \]
    Therefore, if we define $0 < z < \sqrt{\gamma\eps} \frac{\min_{s \in S \backslash \{ 0\}} \|s\|}{\max_{s \in S} \|s\|}$, we have
    \[
    \| x \|^2 + \gamma \eps > \sup_{y \in B(w, \eps)} \|y\|^2
    \]
    for any $x \in K\backslash S_\eps$ such that $\mbox{dist}(x, F_w) \le z$ and any $w \in S\backslash \{0\}$. 
    Using Lemma~\ref{lem:lower_bound_outside_of_Seps}, there exists a constant $\gamma >0$ and a time $T_1$ after which, for any $i \in [n]$, if $x_i(t) \notin S_\eps$, then
    $\|x_i(t+1)\|^2 - \| x_i(t) \|^2 \ge \gamma \eps$. Now, let $t \ge \max(T_1, T_2)$ where $T_2$ is defined in Sublemma~\ref{sublemma:1} for $\lambda = \frac{z}{4}$. Let $i \in [n]$. If $x_i(t) \notin S_\eps$, Lemma \ref{lem:lower_bound_outside_of_Seps} imposes that $\|x_i(t+1)\|^2 - \| x_i(t) \|^2 \ge \gamma \eps$. Since $K$ is bounded, the squared norm of $x_i$ cannot increase indefinitely, hence $x_i(t)$ will reach $B(w, \eps)$ for some $w \in S$ at time $T_3 \ge \max(T_1, T_2)$.
    Then, by the Sublemma~\ref{sublemma:1}, we have $|2\lambda - \dist(x_i(t), F_w)| \xrightarrow[t \rightarrow +\infty]{}0$. Therefore, there exists $T_4 \ge T_3$ such that $\dist(x_i(t), F_w) \le 3 \lambda$, for any $t \ge T_4$. Now, if there exists $T_5 \ge T_4$ such that $x_i(T_5) \notin B(w, \eps)$, since 
    $
    \dist(x_i(T_5), F_w) \le 3\lambda < z$, we know that
    \[
    \| x_i(T_5) \|^2 + \gamma \eps > \sup_{y \in B(w, \eps)} \| y \|^2,
    \]
    and Lemma~\ref{lem:lower_bound_outside_of_Seps} proves that $x_i(t)$ will never re-enter $B(w, \eps)$ at any time $ t \ge T_5$. Therefore, $x_i(t)$ will reach the $\eps$-neighborhood of another point in $S$, say $w'$, to which the same argument applies.
    Since there is only a finite number of elements of $S$, and we have just shown that after a finite time $x_i(t)$ cannot re-enter $\eps$-neighborhood of points in $S$ that it has already visited, this process must stop after a finite time. This means that $x_i$ will eventually enter the $\eps$-neighborhood of an element $w \in S$ and never leave it. As the argument holds for any token $ i $, we have proved that after a finite time, all particles remain in $S_\eps$. 
    
    It remains to prove that particles cannot ``jump'' infinitely often between different connected components of~$S_\varepsilon$. In fact, we know that after $T_2$, we always have $\mathrm{dist}(x_i(t),F_w)\le \varepsilon$, so by Lemma~\ref{lem:max_inner_prod_near_vertices} we can conclude that  $F_{x_i(t)} \subseteq F_w$ for any $t \ge T_2$, where $F_{x_i(t)}$ is the face of $K$ generated by the vertices maximizing the scalar product with $x_i(t)$. This implies that, after $T_2$, if $x_i$ ``jumps'' from $B(w,\varepsilon)$ to $B(w',\varepsilon)$ with $w' \neq w$, then $F_{w'} \subset F_w$. But $w$ is the point that minimizes the norm over $F_w$, so it is unique and the inclusion $F_{w'} \subset F_w$ is strict. Hence, the dimension of the facet $F_{w'}$ is strictly less than that of $F_w$. The same argument applies again if $x_i$ ``jumps'' from $B(w',\varepsilon)$ to another connected component of $S_\varepsilon$. This process must terminate once $x_i$ reaches $B(v,\varepsilon)$ with $v \in V$, since a vertex is a face of $K$ of dimension $0$.
\end{proof}
\begin{remark}
\label{rem:lemmas_valid_for_delta(t)}
    Note that the proof of Theorem~\ref{thm:convergence_localmax_delta(t)} relies on Lemma~\ref{lem:cant_leave_vertices_neigh},Lemma~\ref{lem:diff_max} and Theorem~\ref{thm:S_carac}, where $\delta$ was taken to be a constant. This being said, those results don't rely on $\delta$ being constant and can still be applied in the case of the time-dependent $\delta(t)$.
\end{remark}

\begin{remark}\longthmtitle{Quiescent set around the vertices for vanishing sequences of alignment sensitivity parameters}
\label{rem:case-w-in-V}
    It is worth discussing the special case where $w \in V$. We have that $F_w =F_x= \{w\}$ for any $x \in B(v, \eps)$. Let $\lambda = \frac{\eps}{3}$. Using Sublemma~\ref{sublemma:1}, if $t \ge T_1 := T(\lambda)$ and $x_i(t) \in B(v, \eps)$, then $\big| \frac{2}{3}\eps - \| x_i(t) - w \| \big| \xrightarrow[t \rightarrow +\infty]{}0$. Therefore, there exists $T_2 \ge T_1$ such that $\| x_i(t) - w \| \le \eps$ for any $t \ge T_2$. This proves that there exists a finite time after which the particles cannot leave the $\eps$-neighborhood of any vertices of $K$. This 
    statement is interesting because it shows that $B(w, \eps)$ is a quiescent set even when we replace $\delta$ by $\delta(t)$. 
\end{remark}

\section{Lyapunov stability}
\label{sec:Lyap_stab}
The starting point in this section is to further explore the asymptotic properties of~\eqref{eq:hardmax_delta} using Lyapunov analysis. Our motivation for pursuing this direction is two-fold: first, despite the partial characterizations provided in Theorems~\ref{thm:convex_hull_decreasing} and~\ref{thm:convergence_to_vertices} for the case of a constant alignment sensitivity parameter, a complete asymptotic characterization remains unavailable -- unlike the scenario in Theorem~\ref{thm:convergence_localmax_delta(t)}, where the alignment sensitivity parameter approaches zero. Secondly, at first glance, the dynamics of~\eqref{eq:hardmax_delta} appears to be closely connected to the classical Hegselmann-Krause dynamics, where Lyapunov techniques have been widely used. This naturally raises the question of whether similar analysis can be employed to explain the asymptotic behavior of~\eqref{eq:hardmax_delta}.

The dynamics~\eqref{eq:hardmax_delta} can be reformulated in matrix form as
\begin{equation}
\label{eq:matricial_hardmax_delta}
x(t+1) = \A(t) x(t)\,,
\end{equation}
where $x(t) \in \mathbb{R}^{n \times d}$ is the matrix whose $i$-th row corresponds to the state $x_i(t) \in \real^{1\times d} $ of token $i$, and $\A(t) \in \mathbb{R}^{n \times n}$ is a row-stochastic matrix defined by
\begin{equation}\label{eq:Aij-localmax}
\A_{ij}(t) = \frac{1}{1+\alpha}\delta_{ij} + \frac{\alpha}{1+\alpha} \frac{1}{|C_i^\delta(t)|} \mathbbm{1}_{{j \in C_i^\delta(t)}}\,, 
\end{equation}
where $ C_i^\delta(t) $ is given by~\eqref{eq:neigh-localmax}; note that there is a slight, yet consequential, abuse of notation, as $\A(\cdot)$ is a function of $ x(t) $. 

For the reason that understanding the contrasts with the Hegselmann-Krause (HK) dynamics plays a key role in what follows, we briefly recall a simple version of this classical dynamics: given  $\epsilon >0$, one considers 
\begin{equation}
\label{eq:HK-dyn}
x(t+1) = \HK(t) x(t)\,,
\end{equation}
where $ x(t) \in \real^{n\times d} $ as before, and $ \HK(t) \in \real^{n\times n} $ is given by 
\[
\left(\HK(t)\right)_{ij}=\begin{cases}
\frac{1}{|\N_i(x(t), \epsilon)|} & \text{if } j \in \N_i(x(t)), \\
0 & \text{else},
\end{cases}
\]
with
\begin{equation}\label{eq:Ni}
\N_i(x(t), \epsilon) = \left\{ j \in [n] \;\middle|\; \|x_i(t) - x_j(t)\| \leq \epsilon \right\}.    
\end{equation}

Similar to~\eqref{eq:matricial_hardmax_delta}, the sequence $ (\HK(t))_{t\geq 0} $ is a sequence of row-stochastic matrices. This property is crucially used in the literature to draw conclusion about the convergence properties, and termination time, of the HK-dynamics, see for instance~\cite{SRE-TB-AN-BT:2013}. The key technical idea is to take advantage of the corresponding sequence of \emph{absolute probability vectors}, which we define next; following the literature, from this point on, we sometimes refer to $(A(t))_{t\geq 0} $ as a \emph{chain} of stochastic matrices. 

\begin{definition}
    \label{def:abs_prob_seq}
    Let $(A(t))_{t\geq 0} $ be a chain of stochastic matrices. A sequence of stochastic vectors $(\pi(t))_{t\geq 0}$ is called an \emph{absolute probability sequence} for this chain if for every $t \geq 0$, we have that
    $$
        \pi^T(t) = \pi^T(t+1) \A(t).
    $$
\end{definition}
When $\A(t) = \A$ is a constant matrix, the absolute probability sequence is simply the stationary distribution of $\A$, which always exists. It turns out that the same is true for a chain of row-stochastic matrices, which we state below without proof.
\begin{theorem}[\cite{AK:36}]
    \label{the:existence_abs_prob_seq}
    Every chain of stochastic matrices admits an absolute probability sequence.
\end{theorem}
The reader can refer to~\cite{ES:81} for details on the proof.
Note that the above result holds without any additional ergodicity or irreducibility assumptions. Moreover, the absolute probability sequence is not necessarily unique, unless further ergodicity properties are ensured.

In spite of the large literature on the convergence and ergodicity properties of chains of stochastic matrices, the study of convergence properties of Hegselmann-Krause dynamics, especially in multi-dimensional settings, is far more complex due to the state-dependence of the matrix $\HK(\cdot)$. That said, there has been significant recent progress in understanding the convergence behavior of the HK dynamics (see \cite{ VDB-JMH-JNT:09,SRE-TB-AN-BT:2013,JL:05,LM:05,  AN-BT:2011}). Part of this development relies on Lyapunov studies, which we briefly recall. 
\begin{definition}
    \label{def:lyap1}
    Given a trajectory $(x(t))_{t\geq 0}$ and an absolute probability sequence $ (\pi(t))_{t\geq 0}$ for the chain $(A(t))_{t\geq 0}$, we define the function
    \begin{equation}
        \label{eq:lyap1}
        V(t) = \sum_{i = 1}^n \pi_i(t) \left\| x_i(t) - \pi^T(t) x(t) \right\|^2.
    \end{equation}
\end{definition}

Notably, the one-step decay of this function can be related to the pairwise distance of the states. 
We recall this statement without a proof from~\cite[Theorem 4.3]{BT:12}.

\begin{theorem}
\label{the:estimate_decrease_lyap}
Let $\{A(t)\}$ be a chain of stochastic matrices. Let $\{x(t)\}$ be the solution of \eqref{eq:matricial_hardmax_delta} for some arbitrary initial condition and let $\{\pi(t)\}$ be an absolute probability sequence associated with the chain $\{A(t)\}$. Then, the function~\eqref{eq:lyap1} satisfies 
\begin{equation}
\label{eq:estimate_decrease_lyap}
    V(t) - V(t+1) = \frac{1}{2}\sum_{i,j = 1}^m H_{ij}(t)\lVert x_i(t) - x_j(t)\rVert^2
\end{equation}
with $H_{ij}(t)$ the $ij$-entry of the matrix
\begin{equation}
\label{eq:def_H(t)}
    H(t) = \A'(t) \diag(\pi(t+1))\A(t)
\end{equation}
where $\mathrm{diag}(\pi(t+1))$ is the diagonal matrix whose diagonal is given by the vector $\pi(t+1)$.
\end{theorem}

In the context of the HK-dynamics~\eqref{eq:HK-dyn}, the matrix $ H $ in~\eqref{eq:def_H(t)} is state-dependent. This being said, Theorem~\ref{the:estimate_decrease_lyap} followed by an intricate analysis shows that the HK-dynamics~\eqref{eq:HK-dyn} \emph{terminates in finite time}. As this point, the reader might rightfully wonder as to why such finite-time termination does not hold for the dynamics~\eqref{eq:hardmax_delta}, as shown in Corollary~\ref{cor:no_finite_time_convergence}. Without dwelling into the details, we next highlight two important points related to this. The first one is related to~\eqref{eq:D-mass}: note that the dynamics~\eqref{eq:hardmax_delta} has an added constant diagonal term with fraction of $ \frac{1}{1+\alpha}$ which leads to a geometric rate, and not a finite-time convergence. But besides this, one major added difficulty in analyzing~\eqref{eq:hardmax_delta} in comparison to the HK-dynamics~\eqref{eq:HK-dyn} is lack of ``symmetry'' in the interactions as we describe next.
By definition of the neighborhoods in\eqref{eq:Ni}, for $ i, j \in [n]$, we have that 
\[
i \in \N_j(x(t)) \quad \Leftrightarrow \quad j \in \N_i(x(t)), \quad t \geq 0. 
\]
This ``symmetry'' property in neighborhood structure of particle is key in the analysis of the dynamics; 
in contrast, for~\eqref{eq:hardmax_delta}, the membership $ j \in C_i^\delta(t)$ does not necessarily imply that $ i \in C_j^\delta(t)$ (in fact, we do not even necessarily have $i \in C_i^\delta(t)$). 
In particular, as becomes clear later, the influence cannot solely be captured by the distance of the states of tokens. This lack of symmetry in this sense makes the study of~\eqref{eq:hardmax_delta} more difficult.
In fact, such difficulties already arise in asymmetric variants of the HK-dynamics, including in one-dimensional settings~\cite{AB-MB-BC-HLN:13, JC-DS-BG-BT:15}. It is also well known that when the chain of stochastic matrices is not state-dependent, the analysis becomes significantly more challenging in the presence of unidirectional interactions~\cite{JMH-JNT:12,BT-AN:10-tac}. 

In light of the discussions above, the purpose of our analysis to follow is to investigate the Lyapunov candidate~\eqref{eq:lyap1} in the context of~\eqref{eq:hardmax_delta}. We start by identifying two sets of interest, in the same spirit as~\cite{SRE-TB-AN-BT:2013}.  
\begin{definition}
    \label{def:S1_S2}
    We define the sets:
    \begin{equation*}
    \begin{cases}
        S_1 = \left\{ i \in [n] \;\middle|\; d_i(t) \to 0 \text{ as } t \to +\infty \right\}, \\
        S_2 = \left\{ i \in [n] \;\middle|\; d_i(t) \not\to 0 \text{ as } t \to +\infty \right\},
    \end{cases}
    \end{equation*}
    where $d_i(t) := \max_{k,l \in C_i^\delta(t)} \lVert x_k(t) - x_l(t) \rVert$. 
\end{definition}

We next provide a precise characterization of these two sets. 
\begin{proposition}
    \label{prop:carac_S1_S2}
    The following properties hold:
    \begin{itemize}
        \item[i)] Tokens in  $S_1$ are exactly those converging to the vertices of \( K \), i.e., $ i \in S_1 $ if and only if there exists $ v \in V $ such that $ \|x_i(t) - v\| \to 0$ as $t\to \infty$.

   \item[ii)] There exists a time $T\geq0$ and a positive constant $\gamma>0$ such that
        \begin{itemize}
    \item[a)] 
        $
        S_2 = \{ i \in [n] \; | \; d_i(t) \ge \gamma, \quad \forall t \ge T \};
        $
        \item[b)] For any  $i \in [n]$ , one has $S_1 \cap C_i^\delta(t) \ne \emptyset$  for all \( t \ge T \);
        \item[c)] For any \( i \in S_1 \), one has  $S_2 \cap C_i^\delta(t) = \emptyset$  for all \( t \ge T \).
    \end{itemize}
\end{itemize}    
\end{proposition}
\begin{proof}
We proceed with the proof of each item:
\begin{itemize} 
    \item[i)]Let $0 < \eps \le \delta$ and consider $i \in S_1$. By definition of $S_1$ and Theorem \ref{thm:convex_hull_decreasing}, there exists and $T \ge 0$ such that $d_i(t) \le \frac{\eps}{4}$ and $\eta(t) \le \frac{\eps}{4}$, where $ \eta(t) $ is as defined in~\eqref{eq:eta}, for any $t \ge T$. Let $t\ge T$ and $v$ be such that $\max_{k \in [r]} \innerp{x_i(t)}{v_k} = \innerp{x_i(t)}{v}$.  Suppose that $j$ is such that $x_j(t) \in B(v, \eta(t)) \subset B(v, \frac{\eps}{4})$. Then, applying Lemma \ref{lem:diff_max}, we have :
    \begin{equation*}
        \begin{aligned}
            \max_{\ell \in [n]} \innerp{x_i(t)}{x_\ell(t)} - \innerp{x_i(t)}{x_j(t)} &\le \innerp{x_i(t)}{v - x_j(t)} + \eta(t) \|x_i(t)\| \\
            &\le \eps \|x_i(t)\| \\
            &\le \delta \| x_i(t) \|
        \end{aligned}
    \end{equation*}
    which shows that $ j \in C_i^\delta(t)$. By $d_i(t) \le \frac{\eps}{4}$, it follows $k \in C_i^\delta(t) \implies x_k(t) \in B(v, \frac{\eps}{2})$. This is true for all time $t \ge T$. Therefore, $\dist\left(x_i(t), B(v, \frac{\eps}{2}) \right)\to 0$ as $t\to \infty$. Now, define $\Tilde{T} \ge T$ such that 
    \[
    \dist\left(x_i(t), B(v, \frac{\eps}{2}) \right) \le \frac{\eps}{2}.
    \]
    We just proved that if $i \in S_1$, then for any $0<\eps\le \delta$, there exists a time $\Tilde{T} \ge 0$ such that $\| x_i(t) - v \| \le \eps$. We conclude that $\| x_i(t) - v \| \xrightarrow[t \rightarrow +\infty]{}0$. 
    
    Conversely, suppose that there exists $v \in V$ such that $\| x_i(t) - v \| \rightarrow 0$ as $t\to \infty$. Consider $0 < \eps < \frac{\delta}{2 + 2\alpha}$. Using Sublemma~\ref{sublem:neigh_localized_arnoud_vertices}, we know that after finite time, if $\| x_i(t) -v \| \le \frac{\eps}{4}$, then $\| x_j(t) - v \| \le \frac{\eps}{4}$ for any $j \in C_i^\delta(t)$. Therefore, after finite time, $\|x_i(t) - x_j(t) \| \le \frac{\eps}{2}$ for any $j \in C_i^\delta(t)$ which proves that $d_i(t) \le \eps$. 
    \item[ii(a)] Let $i \in S_2$ and $0 < \eps \le \frac{\delta}{2 + 2\alpha}$. Arguing by contradiction, suppose there exists an increasing sequence of time $(t_p)_{p \in \mathbb{N}}$ diverging to $+\infty$ such that $d_i(t_k) \rightarrow 0$ as $k\to \infty$. Let $k \ge 0$ be large enough so that we can apply the result of Theorem~\ref{thm:convergence_to_vertices} with $\eps$, $d_i(t_k) \le \frac{\eps}{2}$, and $\eta(t_k) \le \frac{\eps}{2}$. Let $v \in V$ be such that $\innerp{x_i(t)}{v} = \max_{k \in [r]} \innerp{x_i(t)}{v_k}$. Following the same reasoning as in i), we know that if $x_j(t_k) \in B(v, \eps)$, then $j \in C_i^\delta(t_k)$. Finally, $d_i(t_k) \le \frac{\eps}{2}$ implies that $x_i(t_k) \in B(v, \eps)$.  By Theorem~\ref{thm:convergence_to_vertices}, we conclude that the state of token $i$ must converge to $v$.  However, this implies that $i \in S_1$, which is impossible. Therefore, for any  $ i \in S_2 $, we have that $d_i(t)\ge\gamma_i $ for some $ \gamma_i>0 $ after some finite time $ \tilde{t}_i$. We now let 
    \begin{equation}\label{eq:gamma}
    \gamma= \min_{i\in S_2} \gamma_i    
    \end{equation}
    and note that $d_i(t)>\gamma $ for all $ i \in S_2$ and $ t \geq \max_i \tilde{t}_i$.
    \item[ii(b)] Let $i \in [n]$, and let $0 < \varepsilon \leq \frac{\delta}{2 + 2\alpha}$. Define $T_1$ as in Theorem~\ref{thm:convergence_to_vertices} with parameter $\varepsilon$, and let $T_2$ be such that $\eta(t) \leq \frac{\varepsilon}{2}$ for all $t \geq T_2$. Set $T = \max(T_1, T_2)$, and fix $t \geq T$. Let $v \in V$ be such that $\langle x_i(t), v \rangle = \max_{k \in [r]}\langle x_i(t), v_k \rangle$. There exists a particle $x_j(t) \in B(v, \eta(t)) \subset B(v, \frac{\varepsilon}{2})$. As in the proofs of items i) and ii(a), it follows that $j \in C_i^\delta(t)$. By Theorem~\ref{thm:convergence_to_vertices}, since $x_j(t) \in B(v, \frac{\varepsilon}{2})$, we conclude that particle $j$ converges to $v$. Finally, using item i), we deduce that $j \in S_1$.
    \item[ii(c)]Let $i \in S_1$. By item i), we know that there exists $v\in S$ such that $\| v - x_i(t) \| \rightarrow0$ as $t\to \infty$. Let $0< \varepsilon \le \frac{\delta}{2 + 2\alpha}$. Let $T_1$ be as defined in Sublemma~\ref{sublem:neigh_localized_arnoud_vertices} for $\eps$, $T_2$ such that $\| v - x_i(t) \| \le \eps$ for any $t\ge T_2$ and $T_3$ as defined in Theorem~\ref{thm:convergence_to_vertices} with $\eps$. Define $T = \max(T_1, T_2, T_3)$ and let $t \ge T$. If $j \in C_i^\delta(t)$, then by Sublemma~\ref{sublem:neigh_localized_arnoud_vertices}, we know that $j \in B(v, \eps)$ and using Theorem~\ref{thm:convergence_to_vertices}, it follows that $\| x_j(t) - v \| \rightarrow0$ as $t\to \infty$ and so $j \in S_1$.    
\end{itemize}    
\end{proof}
\begin{remark}
        \label{rem:S2_influences_S2}
        Note that the fact that $i \in S_1$ is crucial in the proof of item ii(c). Indeed, the statement will not hold if we take $i \in [n] \backslash S_1$. To see this, one can think of two particles that would converge in $S\backslash V$. Those two particles would be in $S_2$ but they would influence each other for time arbitrarily large. This, in fact, has been a major difficulty, as it shows that there exists an underlying dynamics between particles in $S_2$ for which we cannot say anything more than the statement of Proposition~\ref{prop:carac_S1_S2} item ii(c). \oprocend
    \end{remark}
It is important to note that the points~ii(b) and~ii(c) of Proposition~\ref{prop:carac_S1_S2} together imply an asymmetric interaction structure, i.e. after a certain time $T$, tokens in $S_1$ influence the dynamics of tokens in $S_2$, whereas the converse is not true. This is due to the lack of the ``symmetry'' property in neighborhood structure~\eqref{eq:neigh-localmax} highlighted before.

\begin{lemma}
    \label{lem:bound_lyap_1}
    For any $t\ge 0$, we have the following estimate on the decrease of the Lyapunov function~\eqref{eq:lyap1} :
\begin{equation}
    V(t) - V(t+1) \ge \frac{\alpha^2}{(1+\alpha)^2} \frac{1}{2n^2} \sum_{k = 1}^n d_k^2(t) \pi_k(t+1)
\end{equation}
where $d_k(t) = \max_{i,j \in C_k^\delta(t) } \lVert x_i(t) - x_j(t) \rVert$.
\end{lemma}
\begin{proof}
We use Theorem~\ref{the:estimate_decrease_lyap} and evaluate the matrix $H(t)$ defined in \eqref{eq:def_H(t)}:
\begin{equation*}
    H_{ij}(t) = \sum_{k=1}^n \pi_k(t+1) \A_{ki}(t) \A_{kj}(t)\,.
\end{equation*}
Moreover, for fixed $k$,

\begin{equation}
\A_{ki}(t)\A_{kj}(t) = 
\begin{cases}
    \frac{1}{(1+\alpha)^2} + \frac{2\alpha}{(1+\alpha)^2}\frac{1}{|C_k^\delta(t)|} + \frac{\alpha^2}{(1+\alpha)^2}\frac{1}{|C_k^\delta(t)|^2}  & \text{if } i=j=k, \\
    \frac{\alpha}{(1+\alpha)^2} + \frac{\alpha^2}{(1+\alpha)^2}\frac{1}{|C_k^\delta(t)|^2} & \text{if } i=k, j \in C_k^\delta(t), j \neq k, \\
    \frac{\alpha}{(1+\alpha)^2} + \frac{\alpha^2}{(1+\alpha)^2}\frac{1}{|C_k^\delta(t)|^2} & \text{if } j=k, i \in C_k^\delta(t), i \neq k, \\
    0 & \text{if } i \notin C_k^\delta(t) \text{ or } j \notin C_k^\delta(t), \\
    \frac{\alpha^2}{(1+\alpha)^2 |C_k^\delta(t)|^2} & \text{if } i,j \neq k \text{ and } i,j \in C_k^\delta(t).
\end{cases}\label{eq:AijAkl}
\end{equation}
In particular, $\A_{ki}(t)\A_{kj}(t) = 0$ if $i \notin C_k^\delta(t)$ or $j \notin C_k^\delta(t)$, and  
$$
\A_{ki}(t)\A_{kj}(t) \ge \frac{\alpha^2}{(1+\alpha)^2 |C_k^\delta(t)|^2} \quad \text{if } i,j \in C_k^\delta(t)\,.
$$
Therefore,
\begin{equation}
\label{eq:coarse_bound}
    \begin{aligned}
        V(t) - V(t+1) &\ge \frac{1}{2} \frac{\alpha^2}{(1+\alpha)^2}\sum_{k=1}^n \frac{1}{|C_k^\delta(t)|^2} \pi_k(t+1)\sum_{i,j\in C_k^\delta(t)} \lVert x_i(t) - x_j(t) \rVert^2  \\
        &\ge \frac{1}{2n^2} \frac{\alpha^2}{(1+\alpha)^2}\sum_{k=1}^nd_k^2(t) \pi_k(t+1)\,.
    \end{aligned}
\end{equation}
This finishes the proof. 
\end{proof}
\begin{remark}
    \label{rem:def_d_l(t)}
    The reader would be justified in wondering whether it is possible to obtain a better estimate in the second inequality in~\eqref{eq:coarse_bound} in the proof of Lemma~\ref{lem:bound_lyap_1}. Indeed, to get this inequality, we bound $\frac{1}{|C_k^\delta(t)|^2}$ by $\frac{1}{n^2}$ and $\sum_{i,j\in C_k^\delta(t)} \lVert x_i(t) - x_j(t) \rVert^2$ by $d_k^2(t)$. Those two bounds are very coarse. To improve the inequality, one could try to change the definition of $d_k(t)$ by
    \begin{equation}
        \Tilde{d}_k(t) = \frac{1}{|C_k^\delta(t)|}\sqrt{\sum_{i,j \in C_k^\delta(t)} \| x_i(t) - x_j(t) \|^2}\,.
    \end{equation}
    With this definition at hand, the second inequality in~\eqref{eq:coarse_bound} would in fact be an equality, and it would only remain to replace $d_k(t)$ by $\Tilde{d}_k(t)$ in the definition of $S_1$ and $S_2$ to be able to proceed in the exact same way (Lemma~\ref{lem:bound_lyap_2} and Theorem~\ref{the:estimate_decrease_lyap} would still hold). However, the conclusion of Theorem~\ref{the:estimate_decrease_lyap} is that $\pi_k(t+1) = 0$ after a certain time for any $k \in S_2$. Therefore, after finite time, the term $\pi_k(t+1)$ would ``absorb'' all the additional information we obtain by changing the definition of $d_k(t)$, rendering this modification not useful. This observation motivates our choice to keep the original definition that we introduced for $d_k(t)$. \oprocend
\end{remark}
\begin{lemma}
    \label{lem:bound_lyap_2}
    There exists a time $T$ and a constant $\gamma > 0$ such that for any $t \ge T$:
\begin{equation}
    V(t) - V(t+1) \ge \frac{1}{2n^2} \frac{\alpha^2\gamma^2}{(1+\alpha)^2}\sum_{k\in S_2(t)} \pi_k(t+1)\,.
\end{equation}
\end{lemma}
\begin{proof}
The result directly follows from Lemma~\ref{lem:bound_lyap_1} combined with the fact that 
\[
\sum_{k=1}^n \pi_k(t+1) \ge \sum_{k \in S_2(t)} \pi_k(t+1)\,,
\]
since $\pi(t+1)$ is a stochastic vector and $[n] = S_1 \cup S_2$. Finally, we use Proposition~\ref{prop:carac_S1_S2} to obtain the bound $d_i^2(t) \ge \gamma^2$, where $\gamma $ is given by~\eqref{eq:gamma}, for any $i \in S_2$ and $t \ge T$. 
\end{proof}
\begin{definition}
    \label{def:W1_W2}
    For each time $t \geq 0$, define the weights associated with the sets $S_1$ and $S_2$ as
   $$
    W_1(t) := \sum_{k \in S_1} \pi_k(t), \quad \text{and} \quad W_2(t) := \sum_{k \in S_2} \pi_k(t)\,.
   $$
    Since $\pi(t)$ is a probability vector and $S_1 \cup S_2 = [n]$ with $S_1 \cap S_2 = \emptyset$, we have
   $$
    W_1(t) + W_2(t) = 1,
   $$
 and
  $$
    W_1(t), W_2(t) \in [0,1] \quad \text{for all } t \geq 0.
   $$
   \end{definition}

We can now state the main result of the section.

\begin{theorem}
    \label{the:adjoint_of_S2}
    There exists a finite time $T \in \mathbb{N}$ such that $\pi_i(t) = 0$ for all $t \ge T$ and all $i \in S_2$.
\end{theorem}
\begin{proof}
    Suppose, by contradiction, that there exists some $i \in S_2$ such that $\pi_i(t) > 0$ for arbitrarily large times $t$. We choose $T>0$ large enough so that Proposition~\ref{prop:carac_S1_S2}(ii) applies, and pick $t^{*} \ge T$ such that $\pi_i(t^{*}) > 0$. Writing the update relation for the component $\pi_i(t^{*})$, we have
    \begin{equation*}
        \pi_i(t^{*}) = \frac{1}{1+\alpha} \pi_i(t^{*}+1) + \frac{\alpha}{1+\alpha} \sum_{j =1}^n \frac{\mathbbm{1}_{i \in C_j^\delta(t^{*})} }{|C_j^\delta(t^{*})|} \pi_j(t^{*}+1)\,.
    \end{equation*}
   Since the indicator $\mathbbm{1}_{i \in C_j^\delta(t^{*})}$ restricts the sum to those particles $x_j(t^{*})$ influenced by $x_i(t^{*})$, and since by ii(c) in Proposition~\ref{prop:carac_S1_S2}, we have $i \notin C_j^\delta(t^{*})$ for all $j \in S_1$, it follows that
    \begin{equation*}
        \pi_i(t^{*}) = \frac{1}{1+\alpha} \pi_i(t^{*}+1) + \frac{\alpha}{1+\alpha} \sum_{j\in S_2} \frac{\mathbbm{1}_{i \in C_j^\delta(t^{*})} }{|C_j^\delta(t^{*})|} \pi_j(t^{*}+1)\,.
    \end{equation*}
    Summing this equality for $i \in S_2$, we obtain
    \begin{equation*}
        W_2(t^{*}) = \frac{1}{1+\alpha} W_2(t^{*}+1) + \frac{\alpha}{1+\alpha} \sum_{j\in S_2} \frac{1}{|C_j^\delta(t^{*})|} \pi_j(t^{*}+1) \sum_{i \in S_2} \mathbbm{1}_{i \in C_j^\delta(t^{*})}\,.
    \end{equation*}
   
    Since $\sum_{i \in S_2} \mathbbm{1}_{i \in C_j^\delta(t^{*})} \le |C_j^\delta(t^{*})|$, and equality cannot hold because item ii) of Proposition~\ref{prop:carac_S1_S2} guarantees the existence of some particle in $S_1$ influencing $x_j$, it follows that
    \begin{equation*}
        W_2(t^{*}) < \frac{1}{1+\alpha} W_2(t^{*}+1) + \frac{\alpha}{1+\alpha} W_2(t^{*}+1) = W_2(t^{*}+1)\,.
    \end{equation*}
    In particular, $W_2(t) > W_2(t^{*}) > 0$ for any $t \ge t^{*}$. 
    Hence, using Lemma \ref{lem:bound_lyap_2}, regardless of the state of the dynamics, we have
    \begin{equation}
        V(t) - V(t+1) \ge \frac{1}{2n^2} \frac{\alpha^2\gamma^2}{(1+\alpha)^2}W_2(t^{*}) > 0
    \end{equation}
    for any $t \geq t^{*}$. This, however, contradicts the fact that $V(t) \ge 0$ for any $t \ge 0$.
\end{proof}

Some remarks are in order. Since, by Theorem~\ref{the:adjoint_of_S2}, the stochastic vector $\pi(t)$ is eventually concentrated on the set $S_1$ -- in that it eventually has zero components on the set $S_2$ -- the Lyapunov candidate~\eqref{eq:lyap1} does not capture the behavior of the states on the set $S_2$. On the other hand, we already know, by Proposition~\ref{prop:carac_S1_S2}(i), that the tokens in $S_1$ converge to the vertices of the polytope $K$, and vice versa. Therefore, equilibrium points outside the vertices of $K$ -- such as the ones emerging in the example discussed in Remark~\ref{rem:set_S} -- are due to the set $S_2$.
In this sense, Theorem~\ref{the:adjoint_of_S2} demonstrates once more why studying the dynamics of~\eqref{eq:hardmax_delta} is harder in comparison to the classical Hegselmann-Krause dynamics.

\section{Future directions and simulations}\label{sec:future-sim}

We finish this manuscript with outlining some future directions, along with some simulations.

\subsection{Future directions}
The localmax model we have introduced for the attention mechanism aims to more closely approximate the behavior of softmax, while complementing rather than replacing the insights offered by hardmax. As showcased in the paper -- particularly in results such as Theorem~\ref{thm:convergence_to_vertices} -- this comes at the cost of added complexity in the analysis. In light of this, we discuss a necessarily incomplete list of open problems related to the localmax dynamics, followed by more general avenues for future research in the context of attention models for transformers: 

1) In spite of the macro-level results provided in Theorem~\ref{thm:convex_hull_decreasing} concerning convergence to a convex polytope, and the micro-level insights offered by results such as Theorem~\ref{thm:convergence_to_vertices} regarding the behavior of the localmax dynamics near the vertices, a complete understanding of the convergence properties of this dynamics remains elusive. We attribute this difficulty to the asymmetry inherent in the neighborhood structure.

2) Following the previous point, in Section~\ref{sec:Lyap_stab}, we elaborated on why natural approaches commonly used in the opinion dynamics literature fail in the context of localmax dynamics. As we pointed out, the difficulties encountered here stem from the asymmetry in the interaction structure. Indeed, the analysis of Hegselmann-Krause dynamics becomes increasingly challenging when the neighborhood structure is asymmetric, even in one-dimensional settings~\cite{AB-MB-BC-HLN:13,JC-DS-BG-BT:15}. In particular, the Lyapunov methods used in the classical symmetric case no longer apply, although other suitably designed Lyapunov functions -- especially non-monotone ones --may still be viable~\cite{JC-DS-BG-BT:15}. We anticipate that similar approaches could prove fruitful in the study of localmax dynamics.

3) From a control-theoretic perspective, one can formulate the following reachability problem: given a polytope $K$ and a point $w \in \partial K$, determine whether there exists an initial distribution $(x_i(0))_{1 \le i \le n}$ such that:
\begin{itemize}
    \item $\Conv(x_i(t), 1 \le i \le n) \to K$;
    \item $x_i(t) \to w$ for some $i \in [n]$.
\end{itemize}
Lemma~\ref{lem:empty_zone_near_vertices} shows that the answer is negative if $w$ lies within a certain distance of a vertex.  The problem is thus restricted to points in 
$
\partial K \setminus \bigcup_{v \in V} \mathcal{A}_{v}(t).
$
The dependence of this reachability property on the alignment sensitivity parameter $\delta$ remains to be characterized.

    
    4) The convergence of some trajectories to points outside the set $S$ raises the question of whether such points carry any physical or interpretive meaning within the framework of opinion dynamics in the same way the vertices of $K$ do. These points seem to arise only because the threshold $\delta$ is held fixed. It is therefore unclear whether they should be considered meaningful steady states of the system or merely numerical artifacts.
    
Finally, there are many avenues of future research related to all forms of attention models proposed in the literature, as well as in our work. For example, it is of importance to understand what happens if we leave the pure-attention model and add feed-forward layers, or if we include the multi-heads case. Besides this, the idea of considering adding tokens to a system to manipulate the localization of clusters is of great importance in practical applications and worth pursuing.

\subsection{Simulations}

We now present two set of simulations, elaborating further on some of the difficulties that the localmax dynamics imposes, and proposing a related possible direction to tackle these:

\begin{figure}[htb!]
    \centering
    \subfigure[Initial configuration\label{fig:conf_ini_S2_empty}]{
        \includegraphics[width=0.4\textwidth]{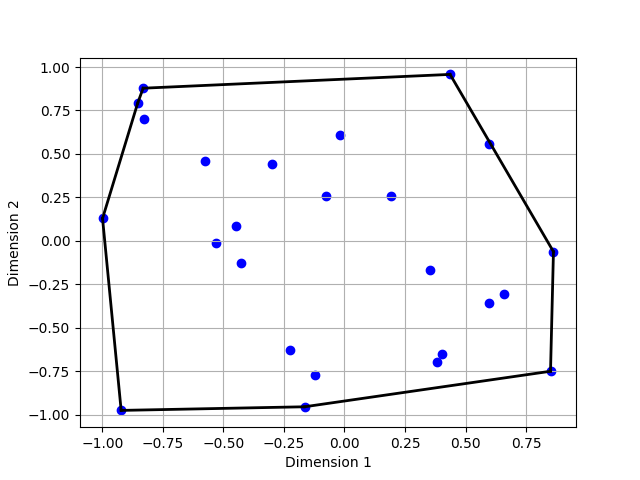}
    }
    \subfigure[Final configuration\label{fig:conf_fin_S2_empty}]{
        \includegraphics[width=0.4\textwidth]{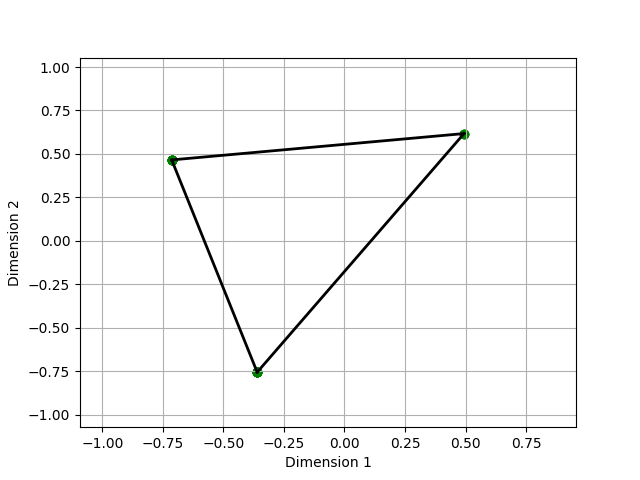}
    }
    \captionsetup{justification=centering}
    \caption{Representation of the initial distribution (Figure \ref{fig:conf_ini_S2_empty}, particles are in blue) and final distribution (Figure \ref{fig:conf_fin_S2_empty}, particles are in green) as well as the convex hull (black lines) for a random initial distribution in the square $[-1, 1]^2$ with parameters $n=25$, $\alpha=0.1$ and $\delta =0.4$. The final points all correspond to vertices of $K$, this shows that $S_2 = \emptyset$.}
    \label{fig:conf_S2_empty}
\end{figure}

In the first simulation, we consider a case where the set $ S_2$ in Definition~\ref{def:S1_S2} is empty. Figure~\ref{fig:conf_S2_empty}(a) and (b), respectively, depict the initial and final convex hull of tokens. As expected, the trajectories of the localmax dynamics converge to the vertices of $ K $ in this case. 
\begin{figure}[htb!]
    \centering
    \subfigure[Initial configuration\label{fig:conf_ini_S2_non_empty}]{
        \includegraphics[width=0.4\textwidth]{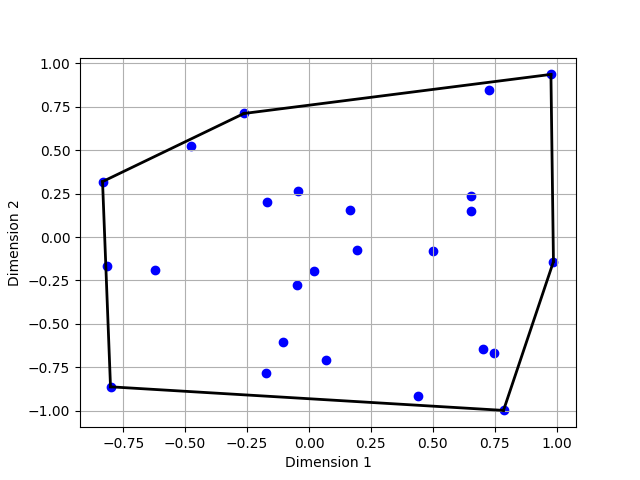}
    }
    \subfigure[Final configuration\label{fig:conf_fin_S2_non_empty}]{
        \includegraphics[width=0.4\textwidth]{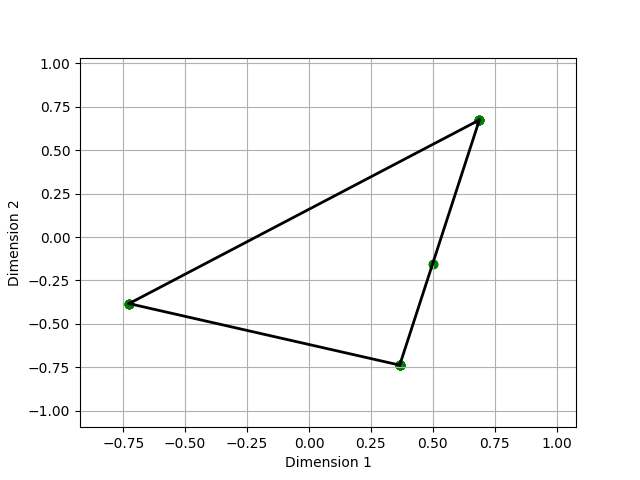}
    }
    \captionsetup{justification=centering}
    \caption{Representation of the initial distribution (Figure \ref{fig:conf_ini_S2_non_empty}, particles are in blue) and final distribution (Figure \ref{fig:conf_fin_S2_non_empty}, particles are in green) as well as the convex hull (black lines) for a random initial distribution in the square $[-1, 1]^2$ with parameters $n=25$, $\alpha=0.1$ and $\delta =0.4$. The point on the left edge in the final configuration that is not a vertex shows that $S_2 \ne \emptyset$.}
    \label{fig:conf_S2_non_empty}
\end{figure}

In the second simulation, we have generated an example where the set $ S_2$ in Definition~\ref{def:S1_S2} is non-empty. Figure~\ref{fig:conf_ini_S2_non_empty}(a) and (b), respectively, depict the initial and final convex hull of tokens. As shown, there is a limit point for the trajectories of the dynamics that is not a vertex of $ K $. This is consistent with our observation that we outlined after the proof of Theorem~\ref{the:adjoint_of_S2}, describing the distinct characteristics of the localmax dynamics that prevents the Lyapunov analysis such as the one for the Hegselmann-Krause dynamics.

\bibliographystyle{plain}  
\bibliography{bib-new}

\end{document}